\documentclass{article}

 \usepackage[main, final]{neurips_2026}

\usepackage[utf8]{inputenc} 
\usepackage[T1]{fontenc}    
\usepackage{hyperref}       
\usepackage{url}            
\usepackage{booktabs}       
\usepackage{amsfonts}       
\usepackage{nicefrac}       
\usepackage{microtype}      
\usepackage{xcolor}         
\usepackage{xspace}
\usepackage{multirow}
\usepackage{graphicx}
\usepackage[table]{xcolor}
\usepackage{graphicx}
\usepackage{subcaption}
\usepackage{wrapfig}
\usepackage{amsmath}
\usepackage{amssymb}
\usepackage{enumitem}
\usepackage{algorithm}
\usepackage{algorithmic}
\usepackage{xcolor}
\usepackage{float}
\usepackage{caption}
\usepackage{enumitem}
\usepackage{tabularx}

\newcommand{\algblock}[1]{%
  \item[] \textcolor{cyan!60!black}{\textbf{#1}}%
}

\newcommand{\algc}[1]{%
  \hfill{\footnotesize\textcolor{cyan!60!black}{$\triangleright$~\textit{#1}}}%
}

\newcolumntype{Y}{>{\raggedright\arraybackslash}X}

\newcommand{\ourmethod}{\textsc{FoCo}\xspace}

\title{ForecastCompass: Guiding Agentic Forecasting with Adaptive Factor Memory}

\author{%
  Yurui Chang, Yongkang Du, Yuanpu Cao, Jinghui Chen, Lu Lin \\
  The Pennsylvania State University \\
  \texttt{\{yuruic, lulin\}@psu.edu}
}

\begin{document}

\maketitle

\begin{abstract}

Agentic forecasting is important for decision-making in dynamic environments, but it remains challenging because agents must reason from incomplete, time-limited evidence and produce calibrated probabilities before outcomes are resolved.
Memory provides a natural mechanism for transferring experience from resolved forecasts to future prediction tasks. However, existing agent-memory methods are not tailored to forecasting, as they typically store past interactions, reflections, or factual associations without explicitly representing reusable predictive factors or calibration knowledge. 
We propose \textbf{Fo}recast\textbf{Co}mpass (\ourmethod), an adaptive factor-based memory framework for agentic forecasting. \ourmethod organizes forecasting experience with a hierarchical forecasting-task taxonomy, enabling retrieval task-relevant forecasting knowledge. It maintains two complementary memory components: factor memory, which captures reusable predictive dimensions, and reasoning memory, which encodes probability updating, uncertainty handling, and calibration principles. Using retrospective analyses as learning signals, \ourmethod iteratively revises memory through a verbalized memory-revision procedure, enabling the agent to accumulate transferable forecasting knowledge over time. Experiments on Prophet Arena and FutureX with GPT-5-mini and Gemini-2.5-Flash show that \ourmethod improves both probabilistic accuracy and calibration. The official code repository for this paper is available at:
\url{https://github.com/yuruic/ForecastCompass}.

\end{abstract}

\section{Introduction}

Large language model based agents have emerged as a promising framework for complex real-world tasks that require information seeking~\citep{nakano2022webgpt}, tool using~\citep{patil2024gorilla,schick2023toolformer}, multi-step reasoning~\cite{yao2023tree,wei2022chain}, and decision-making under uncertainty~\citep{wang2023voyager}. Forecasting is a natural and important instance of this setting, since many high-stakes decisions require anticipating uncertain future events, such as market movements~\citep{box2015time,fama1970efficient}, policy changes~\citep{tetlock2017expert}, and geopolitical developments~\citep{tetlock2016superforecasting}. As LLM-based agents become capable of searching the web, using tools, synthesizing evidence, and reasoning over dynamic information, they provide a promising foundation for agentic forecasting~\citep{yao2022react}.

Building on this perspective, agentic forecasting can be viewed as a more flexible alternative to classical forecasting pipelines~\citep{hyndman2018forecasting,nakano2022webgpt,yao2022react}. Classical forecasting methods in statistics, machine learning, and decision science typically identify the predictive factors, model their temporal dynamics, and update predictions as new observations become available~\citep{ho1998use,hassan2005stock,wu2021dynamic}. However, their modeling process is often constrained by predefined features, structured inputs, and domain-specific assumptions. This limits their flexibility in open-world settings, where relevant evidence may be unstructured, heterogeneous, and distributed across many sources. In contrast, agentic forecasting provides a more flexible paradigm by enabling agents to actively seek information, synthesize diverse evidence, and adapt their reasoning to the forecasting task~\citep{zeng2025futurex,prophet,chang2025survey}. 

Given that forecasting fundamentally depends on identifying and modeling the predictive factors, one natural question is: \emph{can agentic forecasters also be guided to reason in a factor-centric manner?} Instead of relying only on generic experiences or event-level forecast outputs, an agent should learn factor-level knowledge from past events and reuse it for future forecasting. Memory provides a natural mechanism for this by storing factor-granular information that can guide factor selection and probability calibration in the future tasks. 

\begin{wrapfigure}{r}{0.35\textwidth}
    \centering
    \vspace{-10pt}
    \includegraphics[width=\linewidth]{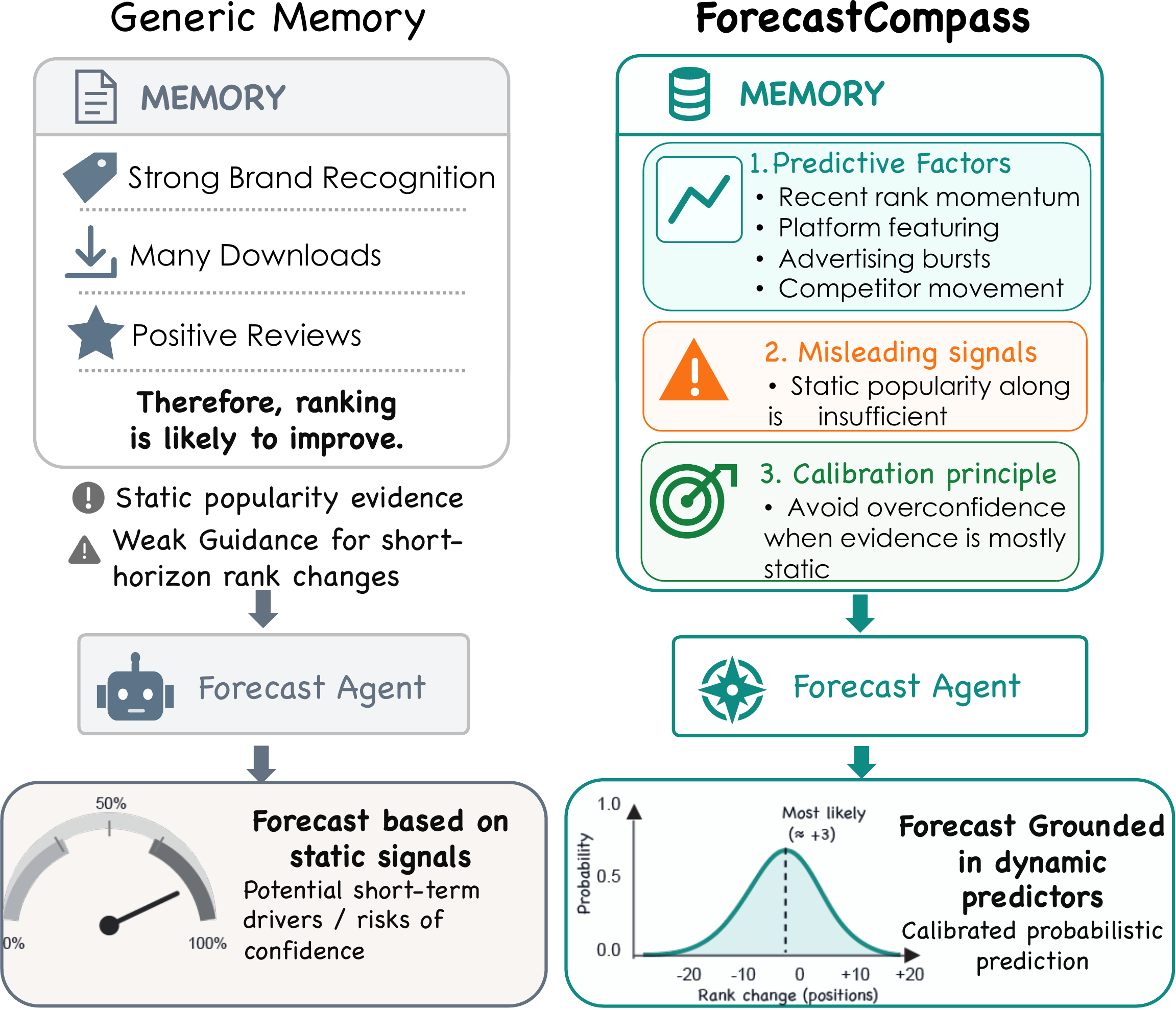}
    \caption{Generic memory v.s.  \ourmethod memory on an example forecasting task.
    }
    \label{fig:example}
    \vspace{-10pt}
\end{wrapfigure}

However, existing memory mechanisms for LLM agents are largely designed for question-answering or problem-solving settings, where memory helps recover an answer, reuse reasoning traces, or avoid past mistakes~\citep{lewis2020retrieval, borgeaud2022improving, chhikara2025mem0}. Forecasting requires memory beyond that used for question answering. 

While QA often relies on retrieving or verifying existing answers, forecasting requires calibrated probabilities for unresolved events under partial and evolving evidence. Moreover, forecasting feedback is noisy and event-specific: a resolved outcome does not directly reveal which predictive factors were missing or misused, and useful lessons may only emerge across multiple trajectories. Forecasting memory must therefore identify recurrent predictive factors, filter misleading or event-specific hindsight, and refine calibration principles. For example, in an app-ranking forecast in Figure~\ref{fig:example}, a generic memory may imply improved ranking based on stored trace such as ``\texttt{the app has strong brand recognition, many downloads, and positive reviews.}'' While useful, this trace does not identify which signals are informative for short-horizon rank changes. A forecasting-specific memory should instead capture dynamic predictive factors such as recent rank momentum, platform featuring, advertising bursts, and competitor movement, while discouraging overconfidence from static popularity evidence. This motivates a \emph{forecasting-specific memory design} that abstracts reusable predictive factors, misleading signals, and calibration principles from retrospective feedback.

To address these challenges, we propose \ourmethod (\textbf{Fo}recast\textbf{Co}mpass), the first forecasting-specific memory framework that organizes past forecasting experience into hierarchical structured memory for future probabilistic prediction. Rather than storing full trajectories or instance-level conclusions, \ourmethod maintains two complementary memory components at the subcategory level: \emph{predictive-factor memory}, which captures reusable signals, misleading evidence patterns, and factor-level lessons, and \emph{calibration-oriented reasoning memory}, which captures how confidence should be adjusted under different evidence conditions. At inference time, the agent retrieves relevant memory to guide evidence search, factor selection, and probability estimation for new unresolved questions. To construct and update this memory, \ourmethod performs contrastive memory revision over resolved forecasting tasks. It contrasts forecasting trajectories with retrospective trajectories, uses their discrepancies as diagnostic signals, and revises the hierarchical memory into reusable predictive factors and calibration principles while avoiding event-level hindsight. Experiments on dynamic forecasting benchmarks show that \ourmethod improves both accuracy and calibration, with learned memory transferring across time periods and model backbones.
Our main contributions are summarized as follows:

\begin{itemize}[leftmargin=*]
\vspace{-2pt}
\item We formulate \emph{factor-centric memory} for agentic forecasting, where memory is organized as subcategory-level predictive factors and calibration-oriented reasoning patterns, rather than event-level trajectories or generic reflections.
\vspace{-2pt}
\item We develop \emph{verbalized factor-memory revision}, a diagnose--aggregate--revise procedure that identifies recurrent lessons across forecasting trajectories and revises memory into transferable predictive-factor and calibration-oriented abstractions while filtering event-specific hindsight.
\vspace{-2pt}
\item We conduct experiments on dynamic forecasting benchmarks including Prophet Arena and FutureX, showing that \ourmethod improves both forecasting accuracy and calibration. We further analyze its transferability across time and models, demonstrating that forecasting memory provides reusable guidance for future unresolved questions.
\end{itemize}

\section{Preliminaries}

\textbf{Agentic forecasting.}
We consider a chronological forecasting setting where a language-model agent
$A_\theta$ makes probabilistic predictions about future events by acquiring
evidence and reasoning over it. For a question $q_t$ with $K_t$ possible
outcomes, the agent observes a time-valid evidence pool $E_t$ and produces
\begin{equation}
(z_t,p_t)=A_\theta(q_t,E_t),
\qquad
p_t\in\Delta^{K_t-1},
\end{equation}
where $z_t$ is the generation trajectory and $p_t$ is the forecast distribution
over the $K_t$ outcomes.

\textbf{Information regimes.}
We distinguish two regimes. In the forecasting regime, the agent can only use
pre-resolution evidence $E_t$ and produces a deployable forecast. After the
outcome is resolved, the retrospective regime provides post-resolution evidence
$E_t^{\mathrm{retro}}$, from which the agent generates a retrospective trajectory
$z_t^{\mathrm{retro}}$. This trajectory is not used for deployment; it serves as
a learning signal that exposes the strengths and failures of the original
forecast.

\begin{figure}[t]
    \centering
    \includegraphics[width=.8\linewidth]{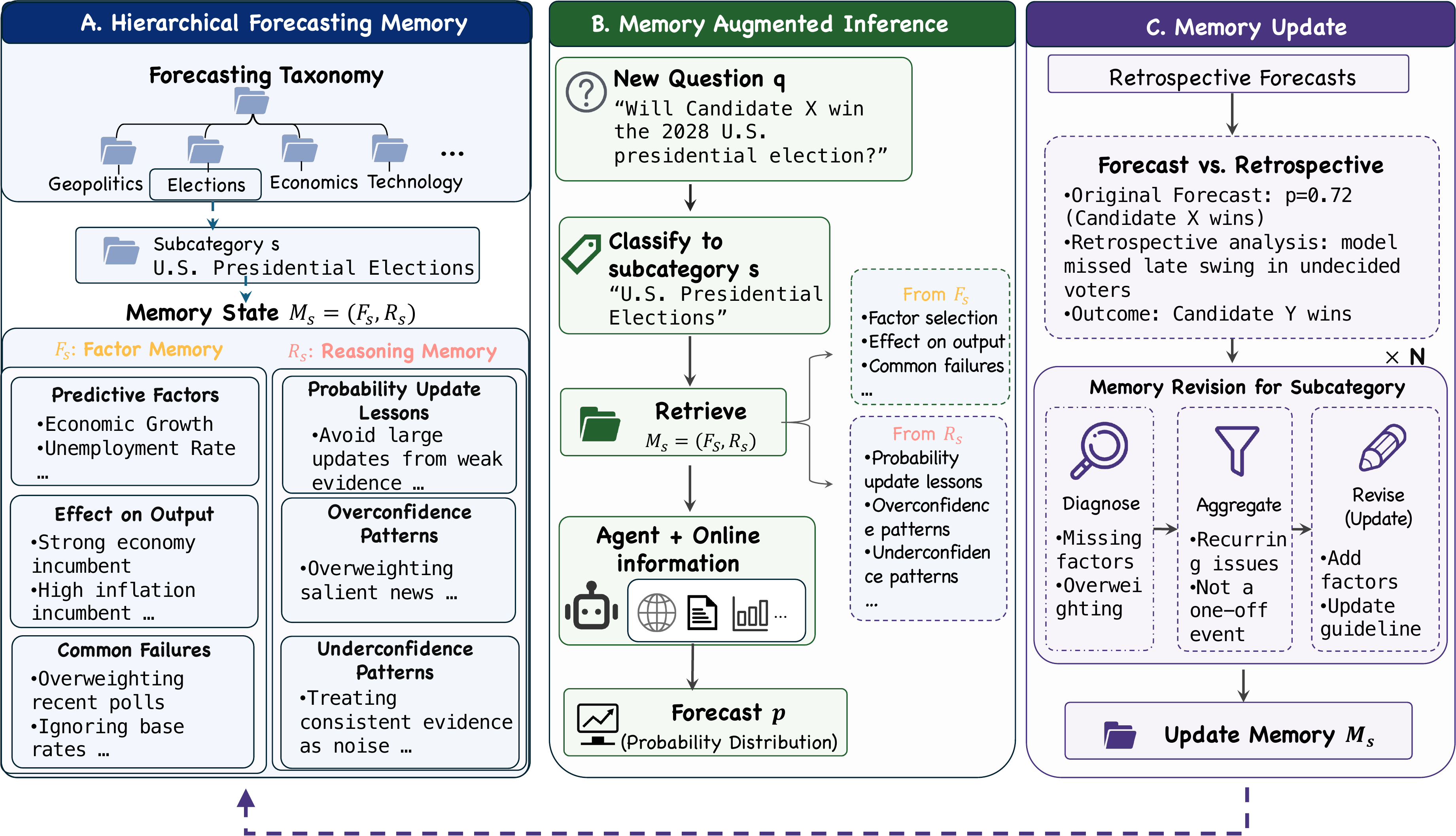}
    \caption{Framework overview. \textit{A}: the hierarchical forecasting memory organizes subcategory-level factor memory and reasoning memory. \textit{B}: a new question is classified into a subcategory, relevant memory is retrieved, and the agent produces a calibrated forecast. \textit{C}: the memory is updated by contrasting original trajectories with retrospective trajectories and iteratively revised.}
    \label{fig:framework}
\end{figure}

\section{Method}
\label{sec:method}

We introduce \ourmethod, a framework that builds forecasting memory from retrospective revisions to predictive factors. \ourmethod organizes memory by a hierarchical taxonomy and subcategory-level states, uses it to guide forecasting, and updates it chronologically as shown in Algorithm~\ref{alg:overall_pipeline}. Detailed algorithms are provided in Appendix~\ref{app:algorithm_details}.

\subsection{Hierarchical Forecasting Memory Representation}
\ourmethod organizes experience with a hierarchical taxonomy. At the update step $w$, the taxonomy $T_w$ consists of a set of categories and fine-grained subcategories. Each forecasting question is associated with one subcategory, and each subcategory serves as the key to a corresponding memory state. For each subcategory $s\in T_w$, \ourmethod maintains a verbal memory state
\begin{equation}
M_{w,s} = (F_{w,s}, R_{w,s})
\end{equation}
where $F_{w,s}$ denotes the predictive-factor memory and $R_{w,s}$ denotes calibration-oriented reasoning memory. 
The predictive-factor memory $F_{w,s}$ is a set of $L_s$ factor entries,
\begin{equation}
F_{w,s} = \{f_{w,s}^{(1)}, \ldots, f_{w,s}^{(L_s)}\},
\end{equation}
where each entry $f_{w,s}^{(i)}$, $i=1,\dots, L_s$ describes a reusable predictive factor for subcategory $s$. Each factor entry contains the factor name, factor-specific evidence checks, common failure modes, and the factor's typical effect on the forecast probability.
These entries guide the agent to identify relevant evidence, avoid misleading uses of individual factors, and estimate how each factor should influence the output distribution. 

The reasoning memory $R_{w,s}$ stores calibration-oriented principles that guide how evidence strength, uncertainty, conflicting signals, and missing information should be translated into probabilistic forecasts. Therefore, $M_{w,s}$ is indexed by the forecasting taxonomy and abstracts resolved forecasting records into reusable subcategory-level forecasting principles rather than event-specific conclusions.

\subsection{Memory-Augmented Forecasting Inference}

For each unresolved question $q_i$ at update step $w$, \ourmethod first assigns it to a category-subcategory pair $(c_i,s_i)$ under the current taxonomy $\mathcal{T}_w$. It then retrieves the corresponding subcategory memory $M_{w,s_i}$ and provides it to the forecasting agent as structured guidance.

Given $q_i$, the accessible evidence pool $E_i$, and the retrieved memory $M_{w,s_i}$, the agent searches or uses tools to gather evidence, conditions its reasoning on the memory, and outputs a forecasting trajectory and probability distribution:
\begin{equation}
(z_i^{\ourmethod}, p_i^{\ourmethod})
=
A_\theta(q_i, E_i, M_{w,s_i}),
\qquad
p_i^{\ourmethod} \in \Delta^{K_i-1}.
\end{equation}

Here, $z_i^{\ourmethod}$ denotes the memory-augmented forecasting trajectory, and $p_i^{\ourmethod}$ denotes the predicted probability distribution over the $K_i$ candidate outcomes.

The retrieved memory guides inference in two complementary ways. The factor memory $F_{w,s_i}$ guides evidence search and factor selection by emphasizing predictive signals and warning against misleading evidence patterns. The reasoning memory $R_{w,s_i}$ guides probability estimation by providing calibration principles for handling uncertainty, conflicting evidence, and incomplete information.

\subsection{Taxonomy and Memory Update Mechanism}

After questions from update step $w$ are resolved, \ourmethod updates the forecasting taxonomy and subcategory-level memories using only past resolved questions, preventing future inference from accessing post-resolution information for unresolved questions.

\paragraph{Taxonomy update.}
Because the taxonomy indexes subcategory memories, it is updated as new forecasting patterns emerge. Let
$T_{w-1}=\{(c,s):c\in\mathcal{C}_{w-1},s\in\mathcal{S}_{w-1,c}\}$
denote the previous taxonomy. Given newly resolved questions $Q_w$, the model assigns each question to an existing category--subcategory pair when possible; otherwise, it places the question in an unmatched pool.

For unmatched questions, the model proposes new category--subcategory pairs, verifies and consolidates them, merges overlapping proposals with existing subcategories, and retains only reusable forecasting patterns. The resulting consolidated candidate set is denoted as
\begin{equation}
\Delta T_w = (\Delta \mathcal{C}_w, \Delta \mathcal{S}_w),
\end{equation}
where $\Delta \mathcal{C}_w$ contains newly introduced categories and $\Delta \mathcal{S}_w$ contains newly introduced subcategories, either under new categories or under existing categories. The taxonomy is then expanded by union:
\begin{equation}
T_w = T_{w-1} \cup \Delta T_w.
\end{equation}

\paragraph{Memory update.}
For each resolved question $q_i \in Q_w$, \ourmethod first routes it to a subcategory $s_i$ under the updated taxonomy $T_w$ and retrieves the corresponding memory $M_{w-1,s_i}$. Only retrieved memories are updated; memories that are not activated by any resolved question remain unchanged. Let $\mathcal{S}_w^{\mathrm{act}}=\{s_i:q_i\in Q_w\}$ denote the set of activated subcategories at update step $w$. For each $s\in\mathcal{S}_w^{\mathrm{act}}$, let $Q_{w,s}$ denote the resolved questions routed to $s$.

After the questions are resolved, \ourmethod constructs retrospective trajectories as learning signals. For each $q_i\in Q_{w,s}$, the agent is given the post-resolution evidence, and generates a retrospective trajectory
\begin{equation}
z_i^{\mathrm{retro}} =
A_\theta^{\mathrm{retro}}(q_i, E_i^{\mathrm{retro}}).
\end{equation}
The retrospective trajectory is not used as a deployable forecast; instead, it explains which evidence and reasoning steps would have supported the realized outcome.

\ourmethod then contrasts the original forecasting trajectory $z_i$ with the retrospective trajectory $z_i^{\mathrm{retro}}$ to obtain a discrepancy signal:
\begin{equation}
\Delta_i = 
\mathrm{Contrast}(z_i, z_i^{\mathrm{retro}}).
\end{equation}
Each discrepancy signal contains two components,
\begin{equation}
\Delta_i = (\Delta_i^F, \Delta_i^R),
\end{equation}
where $\Delta_i^F$ captures factor-level discrepancies, such as missing predictive factors, overused misleading signals, or incorrect factor effects, and $\Delta_i^R$ captures reasoning-level discrepancies, such as miscalibration, overconfidence, underconfidence, or incorrect probability updates. The LLM aggregates these discrepancies into final memory-revision suggestions, which are then applied to revise the corresponding memory state. The finalized memory revision suggestions are:
\begin{equation}
\mathcal{U}_i
=
\left(
\bigcup_{f \in \widetilde{F}_{i,s_i}}
\{\mathrm{Revise}_F(f,\bar{\Delta}_i^F)\},
\;
\mathrm{Revise}_R(R_{w-1,s_i},\bar{\Delta}_i^R)
\right),
\end{equation}

Here, $\widetilde{F}_{i,s_i}$ denotes the subset of factor-memory entries in subcategory $s_i$ that are selected for revision by the aggregated factor-level discrepancy $\bar{\Delta}_i^F$. In other words, \ourmethod does not revise the entire factor memory $F_{w-1,s_i}$; it first identifies the factors that are directly affected by the contrast between the original forecasting trajectory and the retrospective 
trajectory, and only updates this selected subset.

After $N$ iterations of the diagnose--aggregate--revise procedure, the resulting memory is stored as $M_{w,s}$. Forecasting feedback is inherently noisy and event-specific, while reliable predictive factors and calibration patterns are often distributed across multiple resolved forecasts. The iterative procedure is therefore designed to filter spurious or one-off hindsight signals while aggregating recurring evidence into robust, transferable forecasting knowledge. This prevents noisy retrospective information from accumulating in memory and allows useful predictive patterns to be progressively reinforced and refined.

\section{Experiments}
\subsection{Experimental settings}
\label{sec:experimental_settings}

We evaluate \ourmethod with language model based agents on dynamic forecasting benchmarks, where agents reason over predictive factors, gather evidence, and assign calibrated probabilities to future outcome.

\textbf{Models.} 
We use GPT-5-mini~\citep{singh2025openai} and Gemini-2.5-Flash~\citep{comanici2025gemini} as the backbones of the forecasting agent respectively. Unless otherwise specified, memory construction and revision are performed with the same backbone used for forecasting. To test whether the learned memory is tied to a specific model, we also evaluate cross-model transfer by applying memory constructed with Gemini-2.5-Flash to GPT-5-mini. More setup details are provided in Appendix~\ref{app:setup}.

\textbf{Datasets.}
We evaluate \ourmethod on two dynamic forecasting datasets: Prophet Arena~\citep{prophet} and FutureX~\citep{zeng2025futurex}. Both datasets contain time-evolving forecasting questions across a broad range of domains, such as sports, technology, and finance. For each dataset, we select the latest five weeks of data to construct, update, and evaluate memory. This setting reflects the dynamic nature of agentic forecasting, where memory is built from recent resolved questions and revised as new weekly outcomes become available. We follow a chronological evaluation protocol: memory is constructed only from previously resolved questions and is used to forecast future unresolved questions. This prevents post-resolution information from leaking into the current forecast.

\textbf{Evaluation metrics.}
We use Brier score~\citep{glenn1950verification} and expected calibration error (ECE)~\citep{prophet} to evaluate the forecasting performance. Brier score evaluates probabilistic accuracy by measuring the distance between the predicted distribution and the realized outcome. ECE evaluates calibration by comparing predicted confidence with empirical accuracy across confidence bins. Both metrics are lower-is-better. Formal definitions are provided in Appendix~\ref{app:metrics}.

\textbf{Baselines.}
We compare \ourmethod with baselines chosen to cover a broad range of forecasting and memory-design alternatives. 
\textbf{Reference baselines} include \textsc{Base}, which removes external memory, and \textsc{Base} (Retro), a non-deployable retrospective diagnostic reference that analyzes each question after outcome resolution. 
\textbf{Memory-evolution ablations} include \ourmethod{} (Static), which removes weekly memory updates and isolates the value of iterative revision. 
\textbf{Memory-mechanism baselines} are further grouped by the memory mechanism they emphasize: \emph{structured memory}, represented by \textsc{Graphiti}~\citep{rasmussen2025zep}, which stores evolving entity--relation graphs; \emph{retrieval-based long-term memory}, represented by \textsc{Mem0}~\citep{chhikara2025mem0}, which stores and retrieves long-term interaction memories; and \emph{reflective or experience-driven memory}, represented by \textsc{Reflexion}~\citep{shinn2024reflexion} and \textsc{A-Mem}~\citep{xu2025mem}, which construct verbal lessons or agent-experience memories from past trajectories. Together, these baselines provide broad coverage of non-memory, retrospective, static-memory, and representative agent-memory settings.
We provide detailed adaptation protocols in Appendix~\ref{app:baseline_details}, including what information each baseline stores, whether it uses taxonomy information, and how it is updated weekly.

\vspace{-2pt}
\subsection{Main Results}
\label{sec:main_results}

Table~\ref{tab:brier_main_results} reports week-by-week Brier scores on Prophet Arena and FutureX, and Table~\ref{tab:avg_main_results} summarizes the average Brier and ECE scores across all evaluation weeks. Full week-by-week ECE results are provided in Appendix~\ref{app:ece}. The evolution of the taxonomy, with increasing numbers of categories and subcategories as more events are incorporated, is provided in Appendix~\ref{sec:taxonomy_evolution}. We additionally report bootstrap confidence intervals for Brier score in Appendix~\ref{app:bootstrap_ci} to assess robustness of different methods. To further verify that these gains are not specific to a short evaluation window, we extend the FutureX evaluation with six additional weeks (Appendix~\ref{app:more_futurex}), where \ourmethod continues to outperform all baselines in both accuracy and calibration.

\begin{table}[!htbp]
\centering
\small
\resizebox{0.95\linewidth}{!}{%
\begin{tabular}{lcccccccc}
\toprule
\multirow{3}{*}{Method}
& \multicolumn{4}{c}{GPT-5-mini}
& \multicolumn{4}{c}{Gemini-2.5-Flash} \\
\cmidrule(lr){2-5}
\cmidrule(lr){6-9}
& \multicolumn{2}{c}{Prophet Arena}
& \multicolumn{2}{c}{FutureX}
& \multicolumn{2}{c}{Prophet Arena}
& \multicolumn{2}{c}{FutureX} \\
\cmidrule(lr){2-3}
\cmidrule(lr){4-5}
\cmidrule(lr){6-7}
\cmidrule(lr){8-9}
& Brier$\downarrow$ & ECE$\downarrow$
& Brier$\downarrow$ & ECE$\downarrow$
& Brier$\downarrow$ & ECE$\downarrow$
& Brier$\downarrow$ & ECE$\downarrow$ \\
\midrule

\textsc{Base} (Retro)
& 0.109 & 0.079 & 0.197 & 0.209
& 0.187 & 0.098 & 0.241 & 0.279 \\

\textsc{Base}
& 0.150 & 0.114 & 0.241 & 0.263
& 0.202 & 0.106 & 0.266 & 0.299 \\

\midrule

\textsc{Mem0}
& 0.149 & 0.101 & 0.197 & 0.217
& 0.208 & 0.125 & 0.272 & 0.296 \\

\textsc{Reflexion}
& 0.150 & 0.086 & 0.203 & 0.222
& 0.196 & 0.114 & 0.252 & 0.266 \\

\textsc{A-Mem}
& 0.109 & 0.092 & 0.194 & 0.208
& 0.204 & 0.115 & 0.269 & 0.287 \\

\textsc{Graphiti}
& 0.134 & 0.097 & 0.218 & 0.244
& 0.215 & 0.140 & 0.275 & 0.301 \\

\ourmethod{} (Static)
& 0.083 & 0.089 & 0.203 & 0.219
& 0.134 & 0.112 & 0.243 & 0.237 \\

\rowcolor{cyan!8}
\ourmethod{}
& \textbf{0.075} & \textbf{0.077} & \textbf{0.187} & \textbf{0.195}
& \textbf{0.118} & \textbf{0.090} & \textbf{0.216} & \textbf{0.198} \\

\bottomrule
\end{tabular}
}
\vspace{4pt}
\caption{Average Brier and ECE results on Prophet Arena and FutureX across backbone models.}
\label{tab:avg_main_results}
\end{table}

\vspace{-2pt}
\textbf{Factor-centric memory consistently improves forecasting accuracy and calibration.} Across both datasets and backbone models, \ourmethod consistently achieves the best average Brier and ECE scores among all deployable methods. Compared with \textsc{Base}, the gains show that factor-centric memory improves both probabilistic accuracy and calibration, while the improvements over \ourmethod (Static) indicate the importance of iterative memory revision. We include \textsc{Base} (Retro) only as a retrospective diagnostic reference: although it uses post-resolution information, it is not an upper bound because retrospective analysis may induce hindsight bias and overconfident probabilities. Its role is to help diagnose whether errors stem from missing information, reasoning failures, or calibration issues.

\vspace{-2pt}
\textbf{Forecasting-specific memory outperforms general agent-memory baselines.} Existing memory methods such as \textsc{Mem0}, \textsc{Reflexion}, \textsc{A-Mem}, and \textsc{Graphiti} can retain and reuse past experience, but their gains are less consistent across datasets and models. In contrast, \ourmethod obtains the lowest average Brier and ECE scores in all four model-dataset settings in Table~\ref{tab:avg_main_results}. This suggests that simply storing past agent experience is not sufficient for forecasting; organizing memory around reusable predictive factors and calibration-oriented reasoning is more effective. 

\vspace{-2pt}
\textbf{Iterative memory evolution provides additional gains beyond static memory.} \ourmethod{} (Static) improves over \textsc{Base} in several settings, indicating that initialized factor-centric memory is already useful. The full \ourmethod further and consistently improves over this static variant, showing that weekly memory revision refines predictive factors and reasoning patterns as new outcomes become available.

\begin{table*}[h]
\centering
\scriptsize
\resizebox{\textwidth}{!}{%
\begin{tabular}{llcccccccccc}
\toprule
\multirow{2}{*}{Backbone} 
& \multirow{2}{*}{Method}
& \multicolumn{5}{c}{Prophet Arena}
& \multicolumn{5}{c}{FutureX} \\
\cmidrule(lr){3-7}
\cmidrule(lr){8-12}
& & Week 1 & Week 2 & Week 3 & Week 4 & Avg.
  & Week 1 & Week 2 & Week 3 & Week 4 & Avg. \\
\midrule

\multirow{8}{*}{\rotatebox{90}{GPT-5-mini}}
& \textsc{Base} (Retro) & 0.096 & 0.113 & 0.141 & 0.084 & 0.109 & 0.250 & 0.217 & 0.212 & 0.109 & 0.197 \\
& \textsc{Base}            & 0.130 & 0.135 & 0.175 & 0.161 & 0.150 & 0.310 & 0.232 & 0.211 & 0.210 & 0.241 \\
\cmidrule(lr){2-12}
& \textsc{Mem0}            & 0.146 & 0.183 & 0.138 & 0.128 & 0.149 & 0.232 & 0.233 & 0.183 & 0.138 & 0.197 \\
& \textsc{Reflexion}       & 0.130 & 0.147 & 0.155 & 0.167 & 0.150 & 0.246 & 0.183 & 0.213 & 0.170 & 0.203 \\
& \textsc{A-Mem}           & 0.117 & 0.075 & 0.131 & 0.114 & 0.109 & 0.238 & 0.195 & 0.218 & \textbf{0.124} & 0.194 \\
& \textsc{Graphiti}        & 0.128 & 0.135 & 0.137 & 0.134 & 0.134 & 0.250 & 0.233 & 0.218 & 0.172 & 0.218 \\
& \ourmethod{} (Static) & 0.064 & 0.066 & 0.119 & 0.081 & 0.083 & 0.221 & 0.234 & 0.196 & 0.161 & 0.203 \\
\rowcolor{cyan!8}
& \ourmethod{}    & \textbf{0.064} & \textbf{0.069} & \textbf{0.092} & \textbf{0.073} & \textbf{0.075} & \textbf{0.221} & \textbf{0.202} & \textbf{0.180} & 0.144 & \textbf{0.187} \\

\midrule

\multirow{8}{*}{\rotatebox{90}{Gemini-2.5-Flash}}
& \textsc{Base} (Retro) & 0.147 & 0.188 & 0.233 & 0.181 & 0.187 & 0.303 & 0.245 & 0.230 & 0.185 & 0.241 \\
& \textsc{Base}            & 0.169 & 0.226 & 0.221 & 0.190 & 0.202 & 0.329 & 0.272 & 0.237 & 0.227 & 0.266 \\
\cmidrule(lr){2-12}
& \textsc{Mem0}            & 0.180 & 0.225 & 0.221 & 0.205 & 0.208 & 0.319 & 0.281 & 0.246 & 0.242 & 0.272 \\
& \textsc{Reflexion}       & 0.179 & 0.228 & 0.200 & 0.178 & 0.196 & 0.313 & 0.259 & 0.221 & 0.214 & 0.252 \\
& \textsc{A-Mem}           & 0.179 & 0.220 & 0.210 & 0.206 & 0.204 & 0.354 & 0.264 & 0.266 & 0.190 & 0.269 \\
& \textsc{Graphiti}        & 0.175 & 0.208 & 0.248 & 0.230 & 0.215 & 0.354 & 0.261 & 0.243 & 0.242 & 0.275 \\
& \ourmethod{} (Static) & 0.123 & 0.099 & 0.182 & 0.131 & 0.134 & 0.279 & 0.245 & 0.230 & 0.217 & 0.243 \\
\rowcolor{cyan!8}
& \ourmethod{}    & \textbf{0.123} & \textbf{0.115} & \textbf{0.138} & \textbf{0.096} & \textbf{0.118} & \textbf{0.279} & \textbf{0.223} & \textbf{0.217} & \textbf{0.144} & \textbf{0.216} \\

\bottomrule
\end{tabular}
}
\vspace{2pt}
\caption{Brier Score($\downarrow$) results on Prophet Arena and FutureX across different models. }
\label{tab:brier_main_results}
\vspace{-10pt}
\end{table*}

\subsection{Transferability}

We further evaluate two key properties of \ourmethod: the generalizability of factor-centric memory and the importance of iterative memory updating. Across-time transferability tests whether memory constructed from earlier weeks remains useful for later forecasting periods, while across-model transferability tests whether the same memory framework provides consistent benefits across different backbone agents.

\begin{figure}[h]
    \centering

    \begin{subfigure}{0.27\linewidth}
        \centering
        \includegraphics[width=\linewidth]{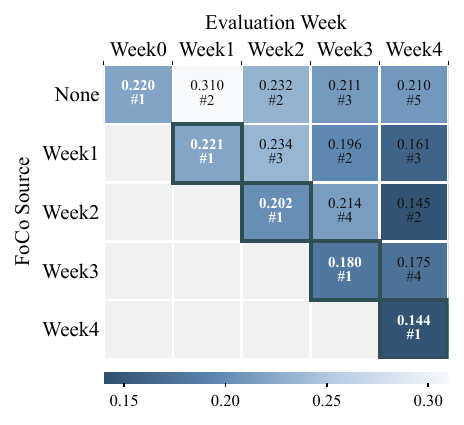}
        \caption{Brier score}
        \label{fig:transfer_brier}
    \end{subfigure}
    \hfill
    \begin{subfigure}{0.27\linewidth}
        \centering
        \includegraphics[width=\linewidth]{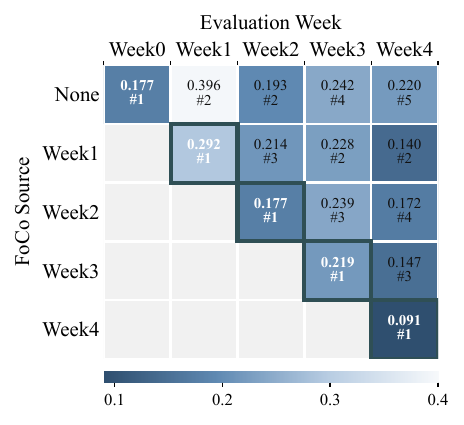}
        \caption{ECE score}
        \label{fig:transfer_ece}
    \end{subfigure}
    \hfill
    \begin{subfigure}{0.4\linewidth}
        \centering
        \includegraphics[width=\linewidth]{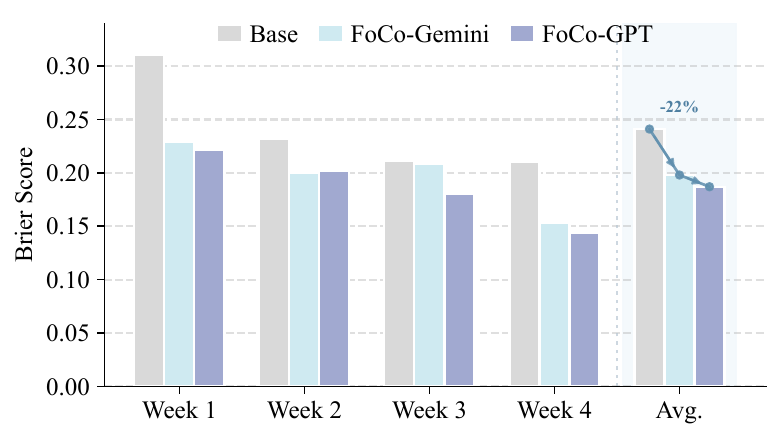}
        \caption{Across-model performance}
        \label{fig:transfer_model}
    \end{subfigure}

    \caption{
    Transferability analysis of \ourmethod on FutureX. 
    We evaluate both across-time transferability and across-model transferability.}
    \label{fig:transfer_all}
    \vspace{-4pt}
\end{figure}

\textbf{Across-time Transferability}
Figure~\ref{fig:transfer_ece} and Figure~\ref{fig:transfer_brier} report across-time transfer results on FutureX. Each row corresponds to the \ourmethod memory source, and each column corresponds to the evaluation week. The ``None'' row denotes the non-memory \textsc{Base} agent, while row Week $k$ denotes using memory constructed up to Week $k$ to forecast later weeks. The left panel reports Brier score, and the right panel reports ECE.

\textbf{Overall, \ourmethod{} provides transferable gains across future weeks.}
Memory improves later-week Brier and ECE scores over the non-memory baseline, with stronger gains generally appearing after it has incorporated more resolved forecasts. Although transfer is not uniformly monotonic across all source--target pairs due to shifts in topics and evidence conditions, the overall pattern suggests that \ourmethod{} transfers both predictive factors and calibration guidance across time, supporting the value of iterative memory evolution.

\textbf{Across-model Transferability}
For across-model transfer, we construct \ourmethod memory with GPT-5-mini and evaluate the same memory with Gemini-2.5-Flash. Figure~\ref{fig:transfer_model} shows that \ourmethod provides consistent gains across different backbone models on Brier scores. Full results including the ECE scores are in the Appendix~\ref{app:more_transferability}. The results indicate that factor-centric memory is not tied to a single backbone agent. Although different models may use the memory differently, the same memory design consistently improves both probabilistic accuracy and calibration. This supports our claim that \ourmethod captures reusable forecasting knowledge, including predictive factors and calibration-oriented reasoning patterns, rather than overfitting to a specific model or time period.

\subsection{Ablation Study}
\vspace{-8pt}
\begin{table}[h]
\centering
\resizebox{\linewidth}{!}{%
\begin{tabular}{lcccccccccc}
\toprule
\multirow{2}{*}{Methods} 
& \multicolumn{2}{c}{Week 1} 
& \multicolumn{2}{c}{Week 2} 
& \multicolumn{2}{c}{Week 3} 
& \multicolumn{2}{c}{Week 4} 
& \multicolumn{2}{c}{Avg.} \\
\cmidrule(lr){2-3}
\cmidrule(lr){4-5}
\cmidrule(lr){6-7}
\cmidrule(lr){8-9}
\cmidrule(lr){10-11}
& Brier$\downarrow$ & ECE$\downarrow$

& Brier$\downarrow$ & ECE$\downarrow$

& Brier$\downarrow$ & ECE$\downarrow$

& Brier$\downarrow$ & ECE$\downarrow$

& Brier$\downarrow$ & ECE$\downarrow$ \\
\midrule
\textsc{Base}              & 0.310 & 0.396 & 0.232 & 0.193 & 0.211 & 0.242 & 0.210 & 0.220 & 0.241 & 0.263 \\
\ourmethod w/o factor     & 0.233 & 0.325 & 0.217 & 0.196 & 0.200 & 0.230 & 0.178 & \textbf{0.083} & 0.207 & 0.209 \\
\ourmethod w/o reasoning  & 0.246 & 0.331 & 0.208 & 0.179 & 0.214 & 0.238 & 0.150 & 0.133 & 0.205 & 0.220 \\
\rowcolor{cyan!8}
\ourmethod                & \textbf{0.221} & \textbf{0.292} & \textbf{0.202} & \textbf{0.177} & \textbf{0.180} & \textbf{0.219} & \textbf{0.144} & 0.091 & \textbf{0.187} & \textbf{0.195} \\
\bottomrule
\end{tabular}
}
\vspace{2pt}
\caption{Ablation study on FutureX with GPT-5-mini.}
\label{tab:ablation}
\vspace{-12pt}
\end{table}

We conduct an ablation study on FutureX with GPT-5-mini to assess the two memory components in \ourmethod. Table~\ref{tab:ablation} compares the full model with two variants: \ourmethod{} w/o factor, which removes predictive-factor memory, and \ourmethod{} w/o reasoning, which removes reasoning and calibration memory. Results show that the two components are complementary. Factor memory improves factor selection by emphasizing predictive signals and reducing reliance on weak or misleading ones, while reasoning memory helps translate factors into calibrated probability estimates. The full model performs best by combining accurate factor selection with calibration-oriented reasoning.

\subsection{Memory Quality Analysis}
\label{mem_quality}
\begin{wrapfigure}{r}{0.4\textwidth}
    \centering
    \vspace{-8pt}
    \includegraphics[width=0.35\textwidth]{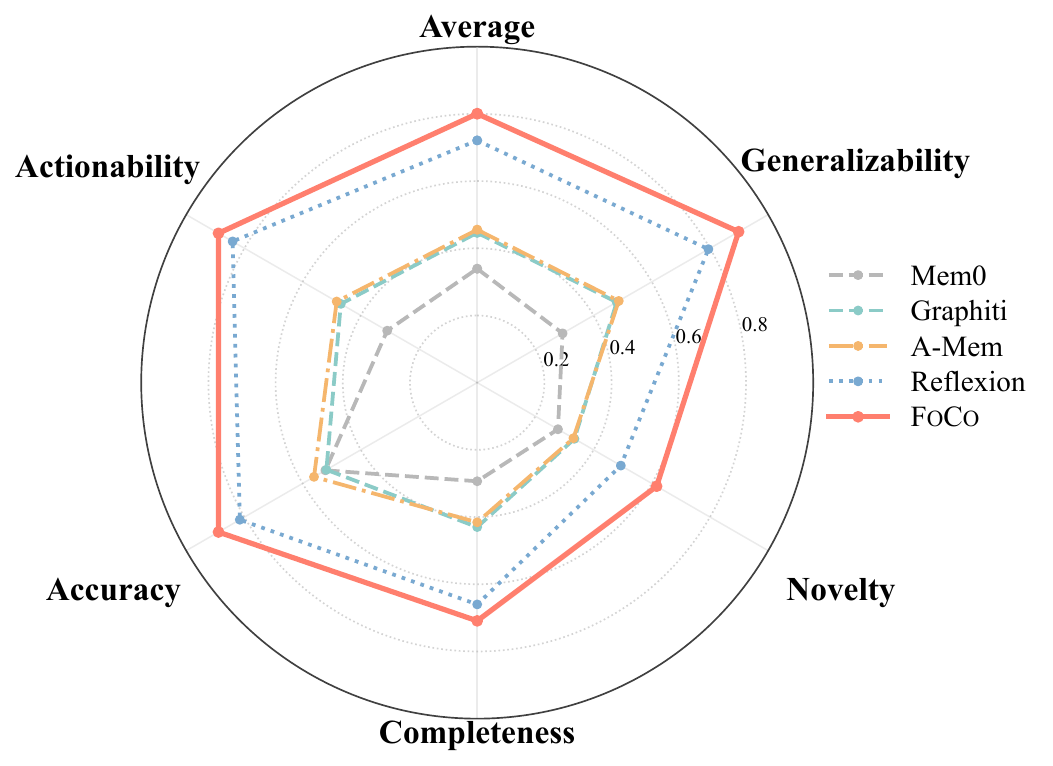}
    \caption{
    Memory quality comparison on Prophet Arena using LLM-as-a-Judge .}
    \label{fig:memory_quality}
\end{wrapfigure}

We use GPT-5.2 to assess whether memory generated by Gemini-2.5-Flash on Prophet-Arena captures reusable forecasting knowledge rather than event-specific details. Memories are scored on five dimensions: generalizability, novelty, completeness, accuracy, and actionability, using the prompt in Appendix~\ref{app:prompt_mem_eval}. As shown in Figure~\ref{fig:memory_quality}, \ourmethod achieves the highest average score and performs strongly across all dimensions. This suggests that its memory abstracts recurring predictive factors and calibration-oriented reasoning patterns, rather than storing event-level forecast records. The strong generalizability and actionability scores further indicate that \ourmethod avoids relying on specific entities, dates, resolved outcomes, or post-resolution facts, supporting its ability to mitigate information leakage through high-level forecasting principles.

\subsection{Case Study}
\begin{figure}[!htp]
    \centering
    \includegraphics[width=\linewidth]{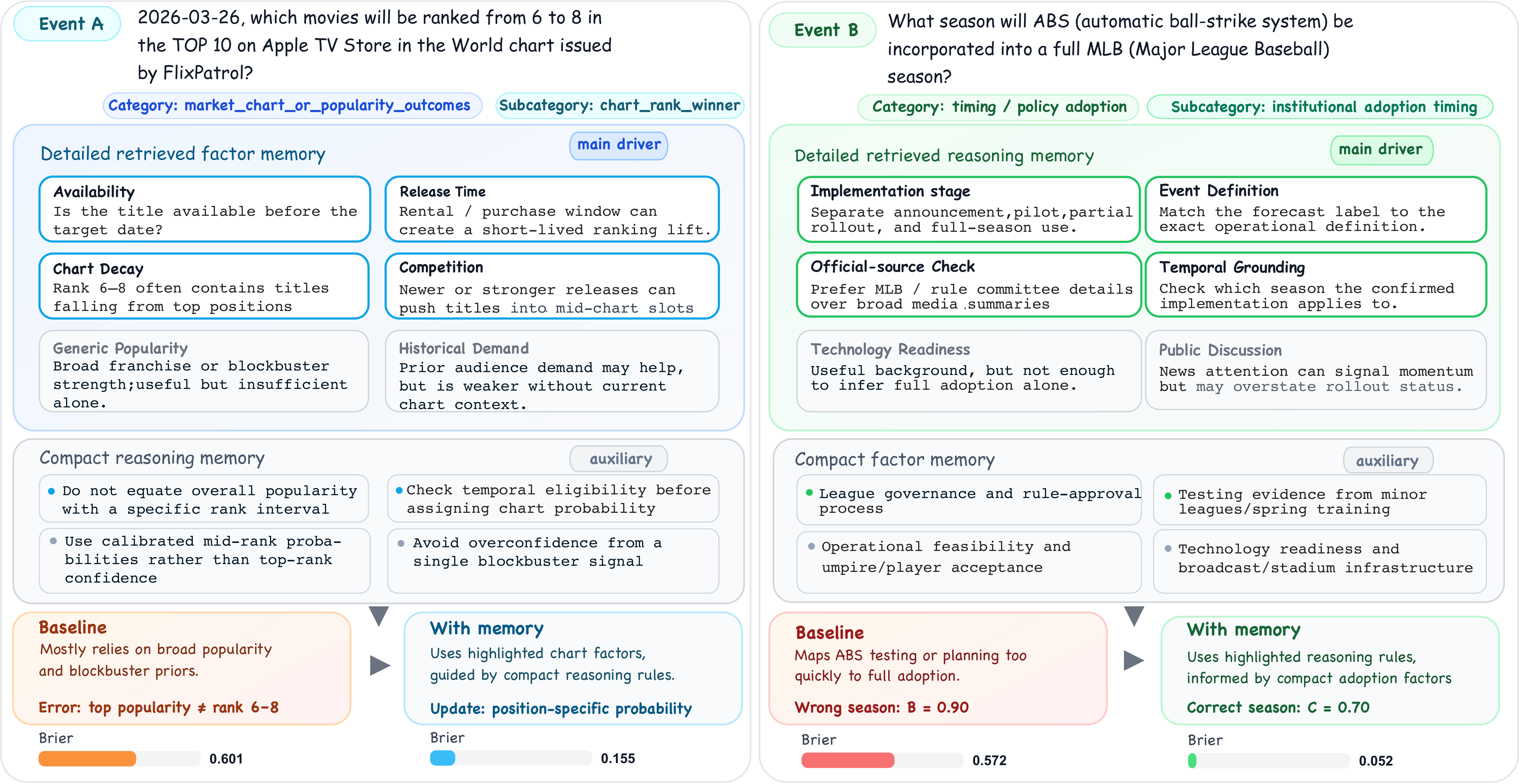}
    \caption{Case Study of how \ourmethod improves forecasts through factor selection and probability calibration.}
    \vspace{4pt}
    \label{fig:case_study}
    \vspace{-15pt}
\end{figure}

Figure~\ref{fig:case_study} illustrates how \ourmethod improves individual forecasts. In Event A, factor memory helps the agent focus on ranking-specific signals, such as availability, release timing, chart decay, and competition, rather than relying mainly on generic popularity. This changes the forecast from a broad popularity judgment to a position-specific probability estimate. In Event B, reasoning memory helps the agent distinguish testing or planning signals from full-season institutional adoption, reducing overconfidence in the wrong season. Across both cases, memory improves the forecast by guiding factor selection and probability calibration, rather than by retrieving event-level answers. We further provide representative examples of the learned memory content in Appendix~\ref{app:mem_visual} and report an LLM-as-a-judge evaluation of memory quality in Section~\ref{mem_quality}.

\section{Related Work}
\vspace{-1em}
\textbf{Forecasting.}
Forecasting aims to make probabilistic predictions about future events under incomplete and evolving evidence~\citep{hyndman2018forecasting, tetlock2016superforecasting}. Classical approaches typically rely on structured historical data, predefined variables, or domain-specific assumptions, such as time-series models~\citep{ho1998use}, probabilistic models~\citep{hassan2005stock,wu2021dynamic}, and expert or crowd forecasting systems~\citep{tetlock2017expert, wolfers2004prediction}. In contrast, language-model-based forecasting agents operate in open-world information environments, where they must gather evidence, synthesize heterogeneous information, identify predictive factors, and reason about uncertain future outcomes~\citep{chang2025survey, cheng2026position, ye2026hindcast}. Existing studies show that LLMs have promising forecasting capabilities, but still face challenges in temporal reasoning, evidence selection, quantitative estimation, and probability calibration~\citep{prophet,zeng2025futurex, yu2026buildingsocialworldmodels, alur2025aiaforecastertechnicalreport}. Our work improves agentic forecasting through memory. Rather than treating each forecasting question independently, \ourmethod uses resolved forecasting tasks to construct reusable predictive-factor and calibration memories for future unresolved questions under chronological constraints.

\textbf{Memory-Augmented Agents.}
Memory has become a key component of language-model agents, enabling them to preserve and reuse information beyond the current context~\citep{zhang2025survey, xu2025mem, chang2026memcollab, zhang2026liveevoonlineevolutionagentic}. Existing memory-augmented agents differ primarily in the content they store. Episodic memory stores past observations, interactions, or trajectories as reusable experience~\citep{park2023generative, packer2023memgpt}; reflection-based memory stores verbal feedback or lessons distilled from prior successes and failures~\citep{shinn2024reflexion, madaan2023self}; skill-based memory stores reusable procedures, strategies, or action programs~\cite{wang2023voyager, cai2023large}; and graph-based memory represents entities, relations, and temporal changes as structured knowledge~\citep{rasmussen2025zep, gutierrez2024hipporag}. At inference time, these memories are typically retrieved or accessed to provide relevant context for new tasks~\citep{zhong2024memorybank,chhikara2025mem0}. However, while existing memory systems preserve past interactions, reflections, or structured knowledge for future reuse, they are not explicitly designed for forecasting, where memory should capture predictive signals, misleading evidence patterns, and calibration principles under incomplete evidence~\citep{tetlock2016superforecasting,guo2017calibration}. \ourmethod addresses this gap by maintaining forecasting-specific memory with two complementary contents: predictive-factor memory and calibration-oriented reasoning memory. It further revises these memories by contrasting forecasting trajectories with retrospective trajectories from resolved tasks.

\section{Conclusion}
This work studies how memory should be designed for agentic forecasting, where agents must make calibrated probabilistic predictions from incomplete and evolving evidence. We show that useful forecasting memory should go beyond storing past trajectories and instead abstract reusable predictive factors and calibration principles. \ourmethod realizes this idea through hierarchical predictive-factor memory and calibration-oriented reasoning memory, revised by contrasting forecasting trajectories with retrospective trajectories from resolved tasks. Experiments on dynamic forecasting benchmarks demonstrate improvements in probabilistic accuracy and calibration, with learned memory transferring across time periods and model backbones. 

\section{Acknowledgements and Disclosures}
The work is supported by the NVIDIA Academic Grant Program. The authors declare no competing interests, including financial relationships outside the submitted work that could be perceived as influencing this research.

\bibliographystyle{plainnat}
\bibliography{reference}

\clearpage
\appendix

\newpage
\section{Additional Experimental Details}

\subsection{Evaluation Metrics}
\label{app:metrics}

We evaluate probabilistic forecasts using both accuracy and calibration metrics. For prediction accuracy, we use the multi-class Brier score:
\[
\mathrm{Brier}(\mathbf{p}_t, \mathbf{y}_t)
=
\sum_{k=1}^{K_t}
(p_{t,k} - y_{t,k})^2,
\]
where $\mathbf{p}_t\in\Delta^{K_t-1}$ is the predicted probability distribution and $\mathbf{y}_t\in\{0,1\}^{K_t}$ is the one-hot encoded realized outcome. Lower Brier score indicates that the predicted probability distribution is closer to the realized outcome.

For calibration, we use expected calibration error (ECE). Given $N$ forecasts, let
\[
\hat{k}_t=\arg\max_k p_{t,k}, 
\qquad
k_t^\star=\arg\max_k y_{t,k},
\qquad
c_t=\max_k p_{t,k},
\]
where $\hat{k}_t$ is the predicted class, $k_t^\star$ is the realized class, and $c_t$ is the prediction confidence. We partition predictions into $B$ confidence bins $\{I_b\}_{b=1}^{B}$. For each bin $I_b$, the empirical accuracy and confidence are defined as
\[
\mathrm{acc}(I_b)
=
\frac{1}{|I_b|}
\sum_{t\in I_b}
\mathbf{1}[\hat{k}_t = k_t^\star],
\]
\[
\mathrm{conf}(I_b)
=
\frac{1}{|I_b|}
\sum_{t\in I_b}
c_t.
\]
The expected calibration error is then
\[
\mathrm{ECE}
=
\sum_{b=1}^{B}
\frac{|I_b|}{N}
\left|
\mathrm{acc}(I_b)
-
\mathrm{conf}(I_b)
\right|.
\]
Lower ECE indicates better calibration, meaning that predicted confidence better matches empirical correctness.

\subsection{Baseline Adaptation Details}
\label{app:baseline_details}

We provide additional details on how each memory baseline is adapted to the forecasting setting. All baselines use the same backbone forecasting agent and follow the same weekly chronological protocol: memory is updated only from previously resolved questions and is used only for future questions. No baseline receives retrospective information about the current evaluation question. The goal of this protocol is to ensure that all methods operate under the same temporal constraint. Also, for all baselines, we use the same factor-centric forecasting prompt in ~\ref{sec:backbone_prompt} as \ourmethod that the agent is first instructed to decompose each question into predictive factors and then produce the final probabilistic forecast.

\paragraph{Reflexion.}
Reflexion stores verbal feedback from past trials and retrieves it for future tasks. For a fair comparison, we use the same factor-centric forecasting prompt as \ourmethod for Reflexion, so the agent first decomposes each question into predictive factors before producing the forecast. For each resolved forecast, we generate a reflection from the original factor-centric forecasting trajectory, the predicted probability distribution, and the ground-truth outcome. The reflection summarizes factor-level successes and failures, such as whether the agent over-relied on weak evidence, missed an important predictive factor, or assigned an overconfident probability. At future forecasting time, relevant reflections are retrieved and prepended to the forecasting prompt. Thus, Reflexion also receives factor-centric outcome feedback from resolved questions, but unlike \ourmethod, it does not organize the resulting knowledge into structured predictive-factor and calibration memories.

\paragraph{Mem0.}
Mem0 is adapted as a general long-term memory system over resolved forecasting records. Each resolved question is treated as an interaction containing the question, forecast reasoning, predicted probability, and realized outcome. Mem0 extracts salient memory snippets from these interactions and retrieves relevant memories for future questions. Unlike FoCo, Mem0 does not maintain a category-subcategory taxonomy and does not explicitly separate predictive factors from reasoning or calibration patterns. Its memory is organized by general relevance retrieval over past interactions.

\paragraph{A-Mem.}
A-Mem is adapted as an experience-level memory baseline. For each resolved forecasting record, we store the question, the agent's reasoning, the predicted probabilities, and the realized outcome as an agent experience. During inference, A-Mem retrieves similar past experiences based on semantic similarity and injects them into the prompt. A-Mem is updated weekly using newly resolved records, following the same chronological protocol as FoCo. However, it does not use the subcategory taxonomy as a core memory structure and does not explicitly aggregate multiple experiences into subcategory-level common factors or calibration rules.

\paragraph{Graphiti.}
Graphiti is adapted as a graph-structured memory baseline. For each resolved forecasting record, an LLM extracts factual triples from the question, reasoning, evidence, and outcome. These triples are inserted into a dynamic graph as entity and relation nodes. During forecasting, relevant graph context is retrieved based on the current question and provided to the agent. Graphiti can partially support cross-question generalization through shared entities or communities, but it does not explicitly construct subcategory-level predictive-factor memory or calibration-oriented reasoning memory.

\section{Additional Experiments}
\label{app:additional_exp}

\subsection{Experimental setup}
\label{app:setup}
For all experiments, we implement the forecasting agent using the OpenAI SDK. Unless otherwise specified, we use GPT-5-mini with medium reasoning effort and set the nucleus sampling parameter top\_p to 0.7. To simulate a realistic forecasting setting and avoid information leakage, we restrict the search window to information available no later than one week before the event resolution time. To account for differences in the number of available events, we set the number of memory-revision epochs to three for Prophet Arena and two for FutureX. The detailed discussion for selecting the epochs is in Appendix~\ref{sec:epoch}.

\subsection{Additional results on more FutureX dataset}
\label{app:more_futurex}

\begin{table*}[htbp]
    \centering
    \resizebox{\textwidth}{!}{%
    \begin{tabular}{l*{7}{rr}}
        \toprule
        & \multicolumn{2}{c}{W5}
        & \multicolumn{2}{c}{W6}
        & \multicolumn{2}{c}{W7}
        & \multicolumn{2}{c}{W8}
        & \multicolumn{2}{c}{W9}
        & \multicolumn{2}{c}{W10}
        & \multicolumn{2}{c}{Average} \\
        \cmidrule(lr){2-3}
        \cmidrule(lr){4-5}
        \cmidrule(lr){6-7}
        \cmidrule(lr){8-9}
        \cmidrule(lr){10-11}
        \cmidrule(lr){12-13}
        \cmidrule(lr){14-15}
        Method
        & Brier & ECE
        & Brier & ECE
        & Brier & ECE
        & Brier & ECE
        & Brier & ECE
        & Brier & ECE
        & Brier & ECE \\
        \midrule
        \textsc{Base} (Retro)
        & 0.215 & 0.184
        & 0.242 & 0.189
        & 0.211 & 0.159
        & 0.206 & 0.177
        & 0.179 & 0.196
        & 0.173 & 0.154
        & 0.204 & 0.177 \\
        \textsc{Base}
        & 0.225 & 0.186
        & 0.225 & 0.253
        & 0.218 & 0.179
        & 0.230 & 0.224
        & 0.224 & 0.216
        & 0.237 & 0.247
        & 0.227 & 0.218 \\
        \midrule
        Reflexion
        & 0.251 & 0.198
        & 0.211 & 0.154
        & 0.214 & 0.151
        & 0.202 & 0.210
        & 0.224 & 0.199
        & 0.218 & 0.216
        & 0.220 & 0.188 \\
        \ourmethod
        & 0.209 & 0.166
        & 0.201 & 0.136
        & 0.198 & 0.119
        & 0.183 & 0.169
        & 0.209 & 0.176
        & 0.178 & 0.198
        & 0.196 & 0.161 \\
        \bottomrule
    \end{tabular}%
    }
    \caption{
        Results on six additional FutureX weeks (W5--W10)
        using GPT-5-mini.
    }
    \label{tab:futurex-additional-weeks}
\end{table*}

To assess whether the observed improvements persist over time, we collected six additional weeks of FutureX data following Week~4 (W5--W10). We evaluated \ourmethod and the three baselines, Base(Retro), Base, and Reflexion, using GPT-5-mini. Table~\ref{tab:futurex-additional-weeks} reports weekly Brier scores and expected calibration errors (ECE), together with their averages across the six weeks.

\ourmethod achieves the lowest average Brier score and ECE among all evaluated methods.  These results demonstrate consistent improvements over Base and Reflexion across the additional weeks, as well as better average performance than all three baselines. We restrict this extended-week analysis to FutureX because it is a live, continuously updated benchmark that had additional resolved weeks available at the time of writing. 

\subsection{Additional results on ECE scores}
\label{app:ece}
\begin{table*}[htbp]
\centering
\scriptsize
\resizebox{\textwidth}{!}{%
\begin{tabular}{llcccccccccc}
\toprule
\multirow{2}{*}{Backbone} 
& \multirow{2}{*}{Method}
& \multicolumn{5}{c}{Prophet Arena}
& \multicolumn{5}{c}{FutureX} \\
\cmidrule(lr){3-7}
\cmidrule(lr){8-12}
& & Week 1 & Week 2 & Week 3 & Week 4 & Avg.
  & Week 1 & Week 2 & Week 3 & Week 4 & Avg. \\
\midrule

\multirow{8}{*}{GPT-5-mini}
& \textsc{Base} (Retro) & 0.035 & 0.131 & 0.075 & 0.073 & 0.079 & 0.309 & 0.183 & 0.234 & 0.109 & 0.209 \\
& \textsc{Base}        & 0.048 & 0.191 & 0.100 & 0.118 & 0.114 & 0.396 & 0.193 & 0.242 & 0.220 & 0.263 \\
\cmidrule(lr){2-12}
& Mem0            & 0.040 & 0.149 & 0.094 & 0.120 & 0.101 & 0.320 & 0.236 & 0.216 & 0.097 & 0.217 \\
& Reflexion       & 0.036 & 0.078 & 0.102 & 0.127 & 0.086 & 0.336 & 0.185 & 0.234 & 0.132 & 0.222 \\
& A-Mem           & 0.052 & 0.103 & 0.103 & 0.111 & 0.092 & 0.291 & 0.180 & 0.262 & 0.100 & 0.208 \\
& Graphiti        & 0.044 & 0.129 & 0.107 & 0.107 & 0.097 & 0.353 & 0.236 & 0.261 & 0.125 & 0.244 \\
& \ourmethod{} (Static) & 0.042 & 0.114 & 0.098 & 0.103 & 0.089 & 0.292 & 0.214 & 0.228 & 0.140 & 0.219 \\
\rowcolor{cyan!8}
& \ourmethod{}    & 0.042 & \textbf{0.064} & 0.101 & 0.099 & \textbf{0.077} & 0.292 & \textbf{0.177} & \textbf{0.219} & \textbf{0.091} & \textbf{0.195} \\

\midrule

\multirow{8}{*}{Gemini-2.5-Flash}
& \textsc{Base} (Retro) & 0.044 & 0.054 & 0.172 & 0.123 & 0.098 & 0.393 & 0.240 & 0.269 & 0.215 & 0.279 \\
& \textsc{Base}            & 0.051 & 0.095 & 0.152 & 0.125 & 0.106 & 0.409 & 0.271 & 0.252 & 0.265 & 0.299 \\
\cmidrule(lr){2-12}
& Mem0            & 0.060 & 0.141 & 0.155 & 0.143 & 0.125 & 0.390 & 0.292 & 0.289 & 0.213 & 0.296 \\
& Reflexion       & 0.062 & 0.161 & 0.116 & 0.117 & 0.114 & 0.363 & 0.246 & 0.231 & 0.223 & 0.266 \\
& A-Mem           & 0.064 & 0.132 & 0.122 & 0.140 & 0.115 & 0.452 & 0.245 & 0.282 & 0.167 & 0.287 \\
& Graphiti        & 0.079 & 0.162 & 0.165 & 0.153 & 0.140 & 0.466 & 0.251 & 0.272 & 0.213 & 0.301 \\
& \ourmethod{} (Static) & 0.068 & 0.092 & 0.152 & 0.135 & 0.112 & 0.232 & 0.230 & 0.286 & 0.200 & 0.237 \\
\rowcolor{cyan!8}
& \ourmethod{}    & 0.068 & 0.119 & \textbf{0.115} & \textbf{0.059} & \textbf{0.090} & \textbf{0.232} & \textbf{0.222} & 0.246 & \textbf{0.091} & \textbf{0.198} \\

\bottomrule
\end{tabular}
}
\caption{ECE Score results on Prophet Arena and FutureX across two backbone models. }
\label{tab:ece_main_results}
\end{table*}
We additionally report ECE results to evaluate the calibration quality of different methods. As shown in Table~\ref{tab:ece_main_results}, \ourmethod consistently achieves the lowest average ECE across both datasets and backbone models, indicating that its forecasting-specific memory not only improves prediction accuracy but also leads to better-calibrated probability estimates. These results further support the effectiveness of structured predictive-factor and calibration-oriented memory for probabilistic forecasting.

\subsection{Additional results on transferability}
\label{app:more_transferability}

\begin{figure}[h]
    \centering

    \begin{subfigure}{0.48\linewidth}
        \centering
        \includegraphics[width=\linewidth]{figures/futurex_transfer_brier_bar.pdf}
        \label{fig:futurex_transfer_brier_full}
    \end{subfigure}
    \hfill
    \begin{subfigure}{0.48\linewidth}
        \centering
        \includegraphics[width=\linewidth]{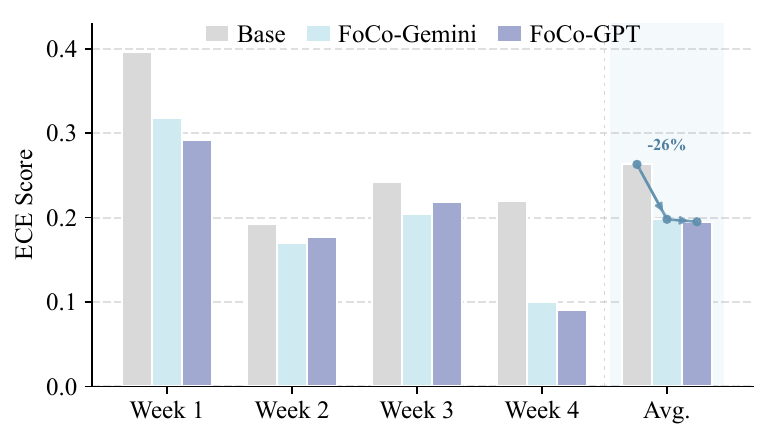}
        \label{fig:futurex_transfer_ece_full}
    \end{subfigure}
    \vspace{-3pt}
    \caption{Across-model memory transfer results on FutureX using GPT-5-mini as the backbone. We compare the no-memory baseline, memory transferred from Gemini-2.5-Flash, and memory from GPT-5-mini.}
    \label{fig:futurex_transfer_two_metrics}
\end{figure}

We further evaluate whether the learned forecasting memory can transfer across backbone models. Figure~\ref{fig:futurex_transfer_two_metrics} reports results on FutureX using GPT-5-mini as the forecasting backbone, comparing the no-memory baseline, memory constructed with Gemini-2.5-Flash, and memory constructed with GPT-5-mini. The transferred memory consistently improves both Brier score and ECE over the no-memory baseline, showing that the learned predictive-factor and calibration-oriented memory captures reusable forecasting knowledge beyond a specific backbone model.

\subsection{Taxonomy Evolution}
\label{sec:taxonomy_evolution}

\begin{figure}[!h]
    \centering
    \vspace{-0.5em}
    \includegraphics[width=0.7\textwidth]{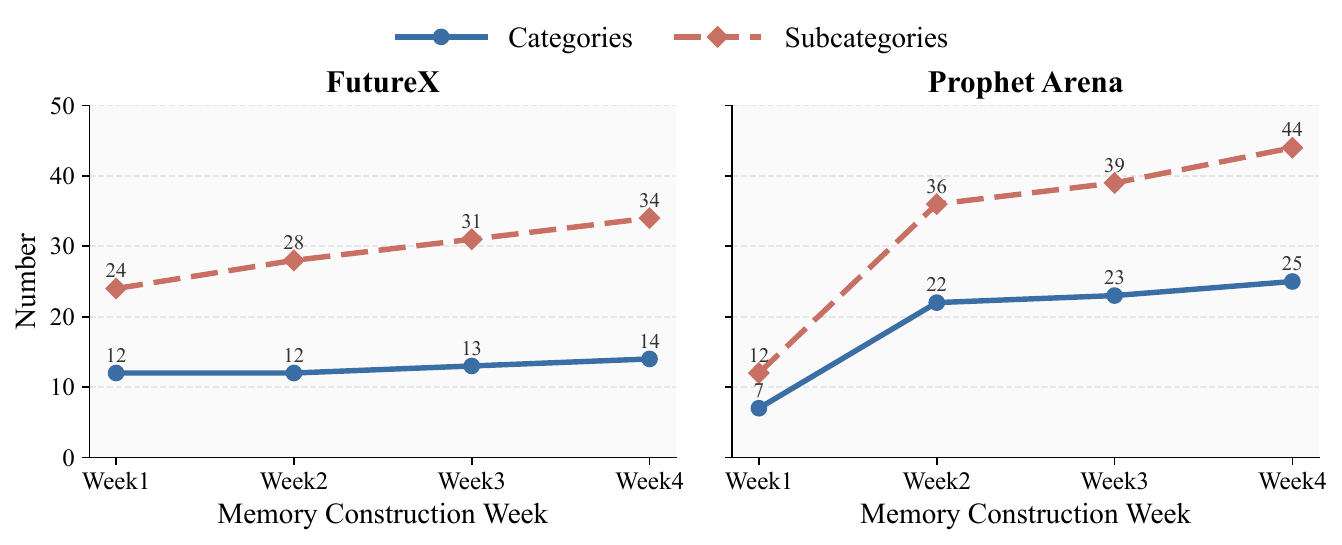}
    \vspace{-0.8em}
    \caption{
    Growth of forecasting taxonomy during memory construction.
    }
    \label{fig:taxonomy_growth}
\end{figure}

Beyond forecasting performance, we analyze how the forecasting taxonomy evolves during memory construction. Figure~\ref{fig:taxonomy_growth} shows the number of categories and subcategories over construction weeks on FutureX and Prophet Arena. For both datasets, the number of categories and subcategories increases as more weekly data is incorporated. This indicates that the forecasting data covers an expanding set of topics over time, and that a fixed taxonomy may be insufficient to organize all emerging questions. The result supports our taxonomy update module, which allows new categories and subcategories to be added as more forecasting questions are observed. It also motivates maintaining memory at the subcategory level, so that newly observed groups of related questions can accumulate their own predictive factors and reasoning patterns.

\textbf{Taxonomy saturation on extended weeks.} To examine whether taxonomy expansion continues indefinitely or converges as a benchmark's topic
space is covered, we extend the taxonomy-evolution analysis to the six additional FutureX weeks
(W5--W10). Table~\ref{tab:taxonomy-extended}
reports the number of categories and subcategories at each week. Category count plateaus at 21, which
suggests that taxonomy growth saturates as recurring forecasting topics are progressively
captured, rather than continuing to expand near-linearly. We attribute this saturation in part to the consolidation step already built into taxonomy
update (Section~\ref{sec:method}), which merges newly proposed subcategories into existing
ones before any new entry is created, rather than treating growth as unconstrained. This
analysis is nonetheless bounded by the benchmark's finite question set and temporal horizon.
Our current mechanism only deduplicates \emph{at creation time}; it does not retroactively
merge or retire existing subcategories that may become redundant or inactive over a much
longer deployment horizon. Extending taxonomy update with such retroactive consolidation is a
natural direction for future work.

\begin{table}[htbp]
\centering
\begin{tabular}{lcccccc}
\toprule
 & Week 5 & Week 6 & Week 7 & Week 8 & Week 9 & Week 10 \\
\midrule
Categories    & 15 & 19 & 19 & 21 & 21 & 21 \\
Subcategories & 37 & 46 & 46 & 51 & 52 & 53 \\
\bottomrule
\end{tabular}
\vspace{2pt}
\caption{Taxonomy size on FutureX (W5--W10), extending the analysis in Figure~\ref{fig:taxonomy_growth}.}
\label{tab:taxonomy-extended}
\end{table}

\subsection{Evaluation of Task Classification}

Prophet Arena and FutureX do not provide ground-truth task categories. We therefore assess routing quality using an independent model, Gemini-2.5-Flash, to judge the subcategory assignments produced by GPT-5-mini. Table below reports the model-judged routing accuracy on Prophet Arena across four weeks.

\begin{table}[hpt]
    \centering
    \renewcommand{\arraystretch}{1.15}
    \setlength{\tabcolsep}{10pt}
    \begin{tabular}{@{}lcccc@{}}
        \toprule
        & \multicolumn{4}{c}{Evaluation period} \\
        \cmidrule(lr){2-5}
        Metric & Week 1 & Week 2 & Week 3 & Week 4 \\
        \midrule
        Accuracy (\%) & 94.1 & 87.7 & 91.8 & 93.0 \\
        \bottomrule
    \end{tabular}
    \vspace{2pt}
    \caption{Classification accuracy on Prophet Arena, judged by
    Gemini-2.5-Flash for assignments made by GPT-5-mini.}
    \label{tab:routing-accuracy}
\end{table}
These results suggest that the routing assignments are largely accurate across the evaluated weeks. However, this evaluation reflects an independent model’s judgments rather than agreement with ground-truth category labels.

\subsection{Bootstrap Confidence Intervals}
\label{app:bootstrap_ci}
To assess the robustness of the forecasting results, we conduct a bootstrap analysis over evaluation questions in Table~\ref{tab:bootstrap_robustness}. For each dataset and method, we resample forecasting questions with replacement and recompute the mean Brier score. We use 10,000 bootstrap resamples and report the original mean Brier score together with the 95\% bootstrap confidence interval, computed from the 2.5th and 97.5th percentiles of the bootstrap distribution. In addition to reporting confidence intervals for each method individually, we conduct a paired bootstrap comparison against the no-memory baseline. For each bootstrap sample, we compute the difference between a method's Brier score and the BASE Brier score on the same resampled set of questions:

\begin{equation}
\Delta_{\text{method}} =
\text{Brier}_{\text{method}} - \text{Brier}_{\text{Base}} .
\end{equation}

\begin{table}[h]
\centering
\small

\resizebox{\linewidth}{!}{

\begin{tabular}{llccc}

\toprule

Dataset & Method & Brier $\downarrow$ & $\Delta$ vs. BASE $\downarrow$ & $p$-value \\

\midrule

\multirow{6}{*}{FutureX}

& \textsc{Base}

& 0.2253 [0.1913, 0.2615]

& --

& -- \\

& \textsc{Mem0}

& 0.2056 [0.1725, 0.2399]

& -0.0197 [-0.0643, 0.0294]

& 0.457 \\

& \textsc{Reflexion}

& 0.2060 [0.1730, 0.2397]

& -0.0193 [-0.0496, 0.0095]

& 0.194 \\

& \textsc{A-MEM}

& 0.2103 [0.1781, 0.2443]

& -0.0150 [-0.0598, 0.0338]

& 0.592 \\

& \textsc{Graphiti}

& 0.2314 [0.1953, 0.2679]

& 0.0061 [-0.0346, 0.0483]

& 0.759 \\

& FOCO

& \textbf{0.1990 [0.1724, 0.2273]}

& \textbf{-0.0263 [-0.0443, -0.0020]}

& \textbf{$<0.001$} \\

\midrule

\multirow{6}{*}{Prophet Arena}

& \textsc{Base}
& 0.1432 [0.1252, 0.1615]
& --
& -- \\
& \textsc{Mem0}
& 0.1537 [0.1359, 0.1720]
& 0.0105 [-0.0130, 0.0364]
& 0.331 \\
& \textsc{Reflexion}
& 0.1435 [0.1255, 0.1627]
& 0.0003 [-0.0219, 0.0210]
& 0.963 \\
& \textsc{A-MEM}
& 0.1066 [0.0899, 0.1244]
& -0.0366 [-0.0599, -0.0101]
& 0.007 \\
& \textsc{Graphiti}
& 0.1349 [0.1166, 0.1548]
& -0.0083 [-0.0314, 0.0179]
& 0.599 \\
\rowcolor{cyan!8}
& \ourmethod
& \textbf{0.0676 [0.0518, 0.0853]}
& \textbf{-0.0756 [-0.0923, -0.0560]}
& \textbf{$<0.001$} \\
\bottomrule
\end{tabular}
}
\vspace{2pt}
\caption{
Bootstrap robustness analysis for Brier score on Prophet Arena and FutureX dataset.
}
\label{tab:bootstrap_robustness}
\end{table}
The bootstrap analysis supports the robustness of the main results.

\subsection{Additional Baseline: Reflexion with Retrospective Trajectory}
\label{app:reflexion_posthoc}
\begin{table*}[h]
\centering
\small
\setlength{\tabcolsep}{3.2pt}
\renewcommand{\arraystretch}{1.12}
\resizebox{\textwidth}{!}{
\begin{tabular}{llcccccccccc}
\toprule
\multirow{2}{*}{Dataset}
& \multirow{2}{*}{Method}
& \multicolumn{2}{c}{Week 1}
& \multicolumn{2}{c}{Week 2}
& \multicolumn{2}{c}{Week 3}
& \multicolumn{2}{c}{Week 4}
& \multicolumn{2}{c}{Avg.} \\
\cmidrule(lr){3-4}
\cmidrule(lr){5-6}
\cmidrule(lr){7-8}
\cmidrule(lr){9-10}
\cmidrule(lr){11-12}
&
& Brier$\downarrow$ & ECE$\downarrow$
& Brier$\downarrow$ & ECE$\downarrow$
& Brier$\downarrow$ & ECE$\downarrow$
& Brier$\downarrow$ & ECE$\downarrow$
& Brier$\downarrow$ & ECE$\downarrow$ \\
\midrule
\multirow{3}{*}{Prophet Arena}
& \textsc{Base}
& 0.130 & 0.048
& 0.135 & 0.191
& 0.175 & 0.100
& 0.161 & 0.118
& 0.150 & 0.114 \\
& \textsc{Reflexion}
& 0.130 & \textbf{0.036}
& 0.147 & 0.078
& 0.155 & 0.102
& 0.167 & 0.127
& 0.150 & 0.086 \\
& \textsc{Reflexion + Retro Trajectory}
& 0.139 & 0.043 
& 0.136 & 0.074
& 0.140	& \textbf{0.097}
& 0.137	& 0.108	
& 0.138	& 0.081 \\
\rowcolor{cyan!8}
& \ourmethod
& \textbf{0.064} & 0.042
& \textbf{0.069} & \textbf{0.064}
& \textbf{0.092} & 0.100
& \textbf{0.073} & \textbf{0.099}
& \textbf{0.075} & \textbf{0.077} \\
\midrule
\multirow{4}{*}{FutureX}
& \textsc{Base}
& 0.310 & 0.396
& 0.232 & 0.193
& 0.211 & 0.242
& 0.210 & 0.220
& 0.241 & 0.263 \\
& \textsc{Reflexion}
& 0.246 & 0.336
& 0.183 & 0.185
& 0.213 & 0.234
& 0.170 & 0.132
& 0.203 & 0.222 \\
& \textsc{Reflexion + Retro Trajectory}
& 0.247 & 0.334
& \textbf{0.173} & \textbf{0.155}
& 0.222 & 0.239
& 0.149 & 0.116
& 0.198 & 0.211 \\
\rowcolor{cyan!8}
& \ourmethod
& \textbf{0.221} & \textbf{0.292}
& 0.202 & 0.177
& \textbf{0.180} & \textbf{0.219}
& \textbf{0.144} & \textbf{0.091}
& \textbf{0.187} & \textbf{0.195} \\
\bottomrule
\end{tabular}
}
\caption{
Additional comparison with a strengthened Reflexion baseline that reflects on post-hoc trajectories.
}
\label{tab:reflexion_posthoc}
\end{table*}

We further examine whether retrospective trajectories provide useful learning signals for memory-based forecasting. To this end, we introduce a strengthened Reflexion baseline, denoted as \textsc{Reflexion + Retro Trajectory}. In contrast to the standard Reflexion baseline, which reflects on the original forecasting trajectory and the resolved outcome, this variant additionally reflects on the retrospective trajectory generated after the outcome is resolved. This comparison tests whether retrospective reasoning can provide useful diagnostic information for future forecasts.

Table~\ref{tab:reflexion_posthoc} shows that retrospective trajectories are indeed valuable. On FutureX, adding retrospective trajectories improves Reflexion from 0.203 to 0.198 in average Brier score and from 0.222 to 0.211 in average ECE. This suggests that retrospective trajectories contain useful information about missing factors, misleading evidence, and calibration errors that can help improve future forecasts.

However, \ourmethod still achieves the best overall performance. This shows that simply storing retrospective reflections is not sufficient. The benefit of \ourmethod comes from converting retrospective diagnostic signals into structured, subcategory-level predictive-factor memory and calibration-oriented reasoning memory. In other words, retrospective trajectories provide valuable raw learning signals, while \ourmethod provides an effective mechanism for abstracting these signals into reusable forecasting knowledge.

\subsection{Cost Analysis}
\begin{table}[hbpt]
\centering
\small
\begin{tabular}{lrrr}
\toprule
Method & \shortstack{Factor\\decomposition}
       & \shortstack{Forecasting\\reasoning} & Total \\
\midrule
\textsc{Base}          & 696 & 3,301 & 3,997 \\
Reflexion     & 690 & 3,067 & 3,757 \\
\ourmethod  & 751 & 3,514 & 4,265 \\
\bottomrule
\end{tabular}
\vspace{4pt}
\caption{Average output-token usage for factor decomposition
and forecasting reasoning. Total denotes the sum across both stages.}
\label{tab:cost_output}
\end{table}
Table~\ref{tab:cost_output} reports the average output-token usage for factor decomposition and forecasting reasoning. \ourmethod achieves substantial forecasting improvements with a modest increase in generated output relative to BASE and Reflexion. The component ablations further support the contribution of both memory types: removing either factor memory or reasoning memory degrades forecasting performance, while the full method performs best. Together, these results demonstrate the effectiveness of combining factor and reasoning memory at a modest additional generation cost.

\subsection{Selection of memory-revision epochs.}
\label{sec:epoch}
\begin{figure}[t]
\centering
\includegraphics[width=0.95\linewidth]{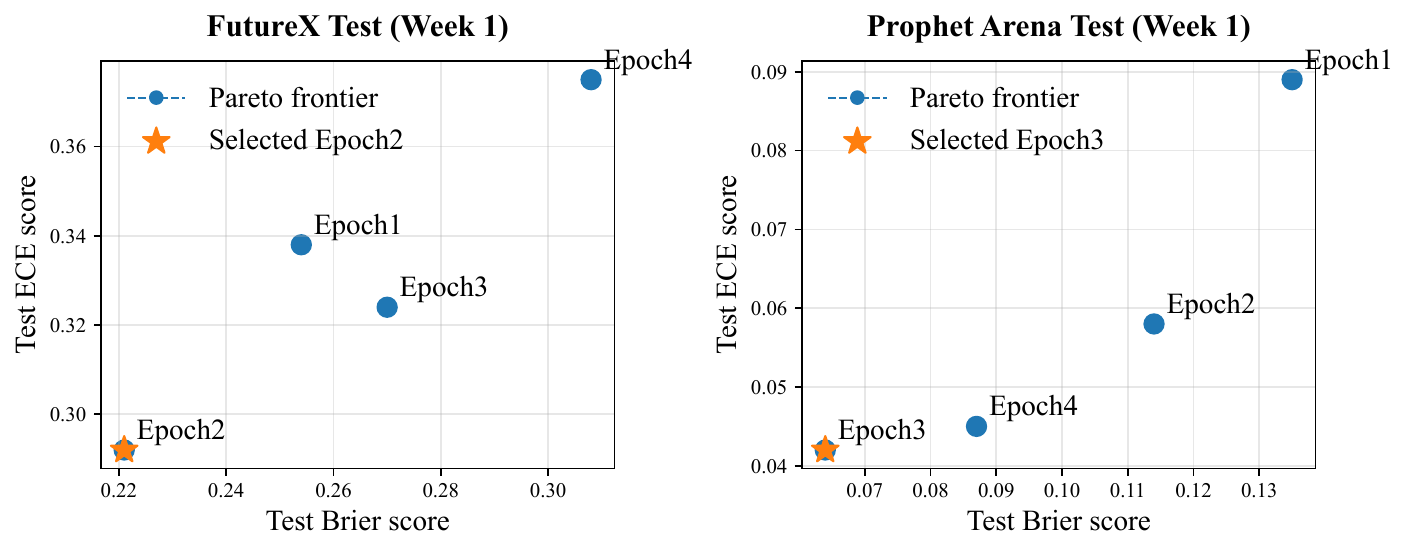}
\vspace{-4pt}
\caption{Brier--ECE trade-off for selecting the number of memory-revision epochs.
Each point denotes one revision epoch.
}
\label{fig:epoch_pareto}
\end{figure}

We adopt a chronological evaluation protocol in which Week 0 is used to initialize memory and Week 1 serves as a validation period for selecting the number of memory-revision epochs. Table~\ref{tab:epoch_selection} reports Brier scores and ECE across different epoch counts on these two weeks. Figure~\ref{fig:epoch_pareto} illustrates the validation Brier--ECE trade-off, with lower values indicating better performance on both metrics. We select two revision epochs for FutureX and three for Prophet Arena, as these settings achieve the lowest validation Brier score and ECE among the evaluated configurations. The selected epoch count is fixed for all subsequent weeks, while memory continues to be updated using questions whose outcomes have become available before each forecast. Week 1 results are reported as validation performance, and subsequent weeks are used to evaluate forecasting performance under the fixed hyperparameter setting.

\begin{table}[t]
\centering
\small
\begin{tabular}{llrrrrrrrr}
\toprule
\multirow{2}{*}{Dataset} & \multirow{2}{*}{Split}
& \multicolumn{2}{c}{Epoch 1}
& \multicolumn{2}{c}{Epoch 2}
& \multicolumn{2}{c}{Epoch 3}
& \multicolumn{2}{c}{Epoch 4} \\
\cmidrule(lr){3-4}
\cmidrule(lr){5-6}
\cmidrule(lr){7-8}
\cmidrule(lr){9-10}
& & Brier & ECE & Brier & ECE & Brier & ECE & Brier & ECE \\
\midrule
FutureX & Train (week 0)
& 0.213 & 0.184 & 0.217 & 0.182 & 0.187 & 0.169 & 0.190 & 0.155 \\
FutureX & Test (week 1)
& 0.254 & 0.338 & 0.221 & 0.292 & 0.270 & 0.324 & 0.308 & 0.375 \\
\midrule
Prophet Arena & Train (week 0)
& 0.114 & 0.113 & 0.076 & 0.111 & 0.079 & 0.100 & 0.079 & 0.103 \\
Prophet Arena & Test (week 1)
& 0.135 & 0.089 & 0.114 & 0.058 & 0.064 & 0.042 & 0.087 & 0.045 \\
\bottomrule
\end{tabular}
\vspace{2pt}
\caption{Effect of memory-revision epochs on development and held-out weeks.}
\label{tab:epoch_selection}
\end{table}

\textbf{Why not validate each memory edit locally.}
An alternative to our aggregation-based revision would be to validate each individual memory edit against a held-out set of resolved questions before accepting it. We find this is not reliable in our chronological setting, where the topic distribution of resolved questions shifts substantially from week to week. Table~\ref{tab:local-validation} reports validation performance on a held-out 20\% split of Week~4 alongside subsequent Week~5 performance, for memories revised with different numbers of epochs: the epoch favored by this local validation (Epoch~1) is not the epoch that performs best on Week~5 (Epoch~2), showing that acceptance decisions from a single-week validation split can overfit to transient topic composition rather than reflecting genuine memory quality. This is consistent with prior memory-augmented agents such as Reflexion~\citep{shinn2024reflexion} and ExpeL~\citep{zhao2024expel}, which likewise revise textual memory without independently validating each edit, relying instead on aggregate downstream performance. \ourmethod mitigates noisy or erroneous updates through aggregation across multiple resolved trajectories and continuous, non-append-only revision, which lets earlier inaccuracies be corrected as new evidence arrives. This is consistent with the absence of accumulated degradation observed across repeated
chronological updates in our experiments.

\begin{table}[htbp]
\centering
\resizebox{\linewidth}{!}{%
\begin{tabular}{lcccc}
\toprule
Revision Epoch & Week 4 Validation Brier$\downarrow$ & Week 4 Validation ECE$\downarrow$ & Week 5 Brier$\downarrow$ & Week 5 ECE$\downarrow$ \\
\midrule
Epoch 1 & \textbf{0.142} & \textbf{0.152} & 0.240 & 0.200 \\
Epoch 2 & 0.168          & \textbf{0.152} & \textbf{0.209} & \textbf{0.166} \\
Epoch 3 & 0.183          & 0.160          & 0.232 & 0.185 \\
\bottomrule
\end{tabular}
}
\vspace{2pt}
\caption{Local validation on a held-out 20\% split of FutureX Week~4, versus subsequent Week~5 performance, across memory-revision epochs.}
\label{tab:local-validation}
\end{table}

\section{Detailed Algorithms}
\label{app:algorithm_details}

This appendix provides the detailed procedures omitted from the main text, including initial memory construction and the internal steps of verbalized factor-memory revision. 
All procedures follow the same chronological constraint as the main pipeline: post-resolution information is used only to revise memory for future forecasts and is never used to modify predictions for the same questions.

\begin{algorithm}[h]
\caption{Initial Memory Construction}
\label{alg:initial_memory_construction}
\small
\begin{algorithmic}[1]
\REQUIRE Initial taxonomy $\mathcal{T}_0$, first-round no-memory forecasting records $\mathcal{R}_0$
\ENSURE Initial memory bank $M_0$

\STATE $\mathcal{T}_0 \leftarrow \mathcal{U}_{\theta}^{\mathrm{tax}}(\mathcal{T}_0,\mathcal{R}_0)$
\algc{initialize and expand taxonomy}

\FOR{each record $r_i\in\mathcal{R}_0$ with question $q_i$}
    \STATE $(c_i,s_i)\leftarrow \mathcal{G}_{\theta}^{\mathrm{tax}}(q_i,\mathcal{T}_0)$
    \algc{assign record to taxonomy}
\ENDFOR

\FOR{each subcategory $s\in\mathcal{S}(\mathcal{T}_0)$}
    \STATE $\mathcal{D}^{\mathrm{res}}_{0,s}\leftarrow\{r_i\in\mathcal{R}_0:s_i=s\}$
    \algc{collect records in subcategory}

    \IF{$\mathcal{D}^{\mathrm{res}}_{0,s}=\emptyset$}
        \STATE \textbf{continue}
    \ENDIF

    \STATE $M_{0,s}\leftarrow \mathcal{I}_{\theta}^{\mathrm{mem}}(\mathcal{D}^{\mathrm{res}}_{0,s})$
    \algc{initialize verbal memory}
\ENDFOR

\RETURN $M_0=\{M_{0,s}:s\in\mathcal{S}(\mathcal{T}_0)\}$
\end{algorithmic}
\end{algorithm}

\begin{algorithm}[h]
\footnotesize
\caption{\ourmethod\ Chronological Update Pipeline}
\label{alg:overall_pipeline}

\noindent\textbf{Input:} Initial records $I_0$, weekly questions
$\{\mathcal{Q}_w\}_{w=1}^{W}$, revision iterations $N$.\\
\noindent\textbf{Output:} Final taxonomy $\mathcal{T}_W$ and memory $M_W$.

\begin{algorithmic}[1]
\STATE $(\mathcal{T}_0, M_0) \leftarrow \mathrm{Initialize}(I_0)$
\algc{initialize taxonomy and subcategory-level memory}

\FOR{$w=1,\ldots,W$}

    \STATE $\mathcal{T}_w \leftarrow
    \mathrm{TaxUpdate}(\mathcal{T}_{w-1}, \mathcal{Q}_w)$
    \algc{assign questions, propose new subcategories, merge taxonomy}
    \FOR{$s \in \mathcal{S}(\mathcal{T}_w)$}

        \STATE $\mathcal{Q}_{w,s} \leftarrow
        \mathrm{Route}(\mathcal{Q}_w, s, \mathcal{T}_w)$
        \algc{select resolved questions in subcategory $s$}
        \STATE $Z_{w,s}^{\mathrm{retro}} \leftarrow
        A_{\theta}(\mathcal{Q}_{w,s}, E^{\text{retro}})$
        \algc{obtain retrospective records and realized outcomes}
        \STATE $M_{w,s} \leftarrow
        \mathrm{MemUpdate}(M_{w-1,s}, \mathcal{Q}_{w,s},
        Z_{w,s}^{\mathrm{retro}}, N)$
        \algc{diagnose, aggregate, and revise memory}

    \ENDFOR

\ENDFOR

\RETURN $\mathcal{T}_W, M_W$
\end{algorithmic}
\vspace{-3pt}
\end{algorithm}

The initialization operator $\mathcal{I}_{\theta}^{\mathrm{mem}}$ summarizes first-round no-memory forecasting records into subcategory-level memory. 
For each subcategory, it constructs two complementary memory components: predictive-factor memory and calibration-oriented reasoning memory. 
The initialized memory abstracts recurring forecasting signals and calibration patterns, rather than storing full trajectories or event-level answers.

Algorithm~\ref{alg:taxonomy_update} gives the detailed procedure for updating the hierarchical forecasting taxonomy. At week $w$, the algorithm starts from the previous taxonomy $\mathcal{T}_{w-1}$ and the incoming question set $\mathcal{Q}_w$. Each question is first classified against the existing taxonomy. If the classifier finds a suitable category-subcategory pair, the question is directly assigned to that pair. Otherwise, the question is marked as unmatched and used to construct a candidate subcategory proposal. All unmatched proposals are collected into a proposal pool. The proposal pool is then verified and merged with the previous taxonomy, where redundant, overly narrow, or semantically overlapping proposals are filtered or merged. This produces the updated taxonomy $\mathcal{T}_w$. Finally, all questions are reassigned under $\mathcal{T}_w$ so that every question can be indexed by a subcategory for memory retrieval and revision.
\begin{algorithm}[t]
\footnotesize
\caption{Taxonomy Update}
\label{alg:taxonomy_update}
\begin{flushleft}
\textbf{Input:} Previous taxonomy $\mathcal{T}_{w-1}$, incoming questions $\mathcal{Q}_w$.\\
\textbf{Output:} Updated taxonomy $\mathcal{T}_w$ and assignments $\{(c_i,s_i)\}_{q_i\in\mathcal{Q}_w}$.
\end{flushleft}

\vspace{-0.8em}

\begin{algorithmic}[1]
\STATE $\mathcal{P}_w \leftarrow \emptyset$

\FOR{$q_i \in \mathcal{Q}_w$}
    \STATE $o_i \leftarrow \mathcal{G}_{\theta}^{\mathrm{tax}}(q_i,\mathcal{T}_{w-1})$
    \algc{match question to existing taxonomy}
    \IF{$o_i=\emptyset$}
        \STATE $\tilde{s}_i \leftarrow
        \mathrm{ProposeSubcategory}_{\theta}(q_i,\mathcal{T}_{w-1})$
        \STATE $\mathcal{P}_w \leftarrow \mathcal{P}_w \cup \{(q_i,\tilde{s}_i)\}$
        \algc{unmatched question induces a proposal}
    \ENDIF
\ENDFOR

\STATE $\mathcal{T}_w \leftarrow
\mathcal{U}_{\theta}^{\mathrm{tax}}(\mathcal{T}_{w-1},\mathcal{P}_w)$
\algc{merge proposal pool into taxonomy}

\FOR{$q_i \in \mathcal{Q}_w$}
    \STATE $(c_i,s_i) \leftarrow
    \mathcal{G}_{\theta}^{\mathrm{tax}}(q_i,\mathcal{T}_w)$
\ENDFOR

\RETURN $\mathcal{T}_w,\{(c_i,s_i)\}_{q_i\in\mathcal{Q}_w}$
\end{algorithmic}
\end{algorithm}

\begin{algorithm}[t]
\caption{Detailed Memory Construction and Revision}
\label{alg:detailed_memory_construction}
\small
\begin{algorithmic}[1]
\REQUIRE Taxonomy $\mathcal{T}_w$, previous memory bank $M_{w-1}$, weekly questions $\mathcal{Q}_w$, forecast records $\{r_i^{\mathrm{fore}}\}$, resolved outcomes $\{y_i\}$, revision iterations $N$
\ENSURE Updated memory bank $M_w$

\algblock{Retrospective record construction}
\FOR{each resolved question $q_i \in \mathcal{Q}_w$}
    \STATE Generate retrospective record $r_i^{\mathrm{retro}}$
    \algc{uses post-resolution evidence}
    \STATE $(c_i,s_i)\leftarrow \mathcal{G}_{\theta}^{\mathrm{tax}}(q_i,\mathcal{T}_w)$
    \algc{recover taxonomy assignment}
\ENDFOR

\algblock{Subcategory grouping}
\FOR{each subcategory $s\in\mathcal{S}(\mathcal{T}_w)$}
    \STATE $I^{\mathrm{res}}_{w,s}
    \leftarrow
    \{(q_i,r_i^{\mathrm{fore}},r_i^{\mathrm{retro}},y_i):q_i\in\mathcal{Q}_w,\ s_i=s\}$
    \algc{resolved records for subcategory}

    \algblock{Verbalized factor-memory revision}
    \STATE $M_{w,s}^{(0)}\leftarrow M_{w-1,s}$

    \FOR{$n=0,\ldots,N-1$}
        \STATE $\Delta_{w,s}^{(n)}
        \leftarrow
        \mathcal{D}_{\theta}^{\mathrm{mem}}
        (M_{w,s}^{(n)},I^{\mathrm{res}}_{w,s})$
        \algc{diagnose missing factors and calibration errors}

        \STATE $\bar{\Delta}_{w,s}^{(n)}
        \leftarrow
        \mathcal{A}_{\theta}^{\mathrm{mem}}
        (\Delta_{w,s}^{(n)})$
        \algc{aggregate event-level diagnostics}

        \STATE $M_{w,s}^{(n+1)}
        \leftarrow
        \mathcal{R}_{\theta}^{\mathrm{mem}}
        (M_{w,s}^{(n)},\bar{\Delta}_{w,s}^{(n)})$
        \algc{revise verbal memory}
    \ENDFOR

    \STATE $M_{w,s}\leftarrow M_{w,s}^{(N)}$
\ENDFOR

\RETURN $M_w=\{M_{w,s}:s\in\mathcal{S}(\mathcal{T}_w)\}$
\end{algorithmic}
\end{algorithm}

The diagnosis operator $\mathcal{D}_{\theta}^{\mathrm{mem}}$ compares the original forecast record, the retrospective record, and the realized outcome to identify missing predictive factors, misleading evidence, and calibration errors. 
The aggregation operator $\mathcal{A}_{\theta}^{\mathrm{mem}}$ compresses event-level diagnostics into recurring subcategory-level patterns. 
The revision operator $\mathcal{R}_{\theta}^{\mathrm{mem}}$ rewrites the memory state into updated predictive-factor memory and calibration-oriented reasoning memory.

The revised memory is constrained to encode transferable forecasting knowledge rather than instance-level answers. 
Specifically, it should retain recurring predictive signals, factor interactions, failure modes, and confidence-adjustment principles, while excluding copied post-resolution evidence, resolved outcomes, exact event dates, final rankings, and entity-specific conclusions. 
This constraint allows retrospective records to serve as learning signals for future questions without turning memory into an event-level answer cache.

\section{Forecasting Memory Construction Details}
\label{app:visual}
\subsection{Representative subcategory memories}
\label{app:mem_visual}
We provide two representative subcategory-level memories learned by \ourmethod. Each memory contains two components: \emph{factor memory}, which stores reusable predictive dimensions and their typical effects on forecasts, and \emph{reasoning memory}, which stores calibration, probability-update, and uncertainty-handling lessons.
\begin{table*}[t]
\centering
\footnotesize
\setlength{\tabcolsep}{4pt}
\renewcommand{\arraystretch}{1.03}

\begin{tabularx}{\textwidth}{p{0.27\textwidth} Y}
\toprule
\multicolumn{2}{l}{\textbf{Representative Memory Example 1: \texttt{team\_sports\_match\_winner}}} \\
\midrule
\textbf{Factor Memory} & \textbf{Typical Effect on Output} \\
\midrule

Market-implied expectation
& Anchors forecasts close to market consensus for liquid events; as liquidity declines or independent model signals strengthen, forecasts move away from market priors with proportional uncertainty. \\

Event-state and canonical instance mapping
& Determines whether the task remains probabilistic or should collapse toward a recorded outcome; ambiguous or resumed instances require branching and retained uncertainty. \\

Provenance strength, independence, and corroboration count
& High-tier, independently corroborated primary sources justify large probability shifts and smaller residuals; single secondary or aggregated reports produce only modest updates. \\

Evidence freshness, retrievability, and cutoff admissibility
& Fresh, provably pre-cutoff primary records produce the strongest updates; post-cutoff or archival-only evidence should not drive pre-cutoff collapse. \\

Personnel availability, role leverage, and replacement depth
& Verified availability or absence of high-leverage personnel often produces the largest conditional shifts; uncertain availability should be scenario-branched. \\

Venue, travel burdens, and tactical matchup fit
& Produces moderate adjustments for home side, travel, venue micro-effects, or tactical fit; strong mismatches can materially alter expectations. \\

Base rates, competition stage, structural tier, and short-run form
& Stabilizes priors near long-run expectations absent strong event-specific evidence; structural tier gaps set upset floors and regularized short-run signals shift posteriors modestly. \\

Outcome-space completeness and residual finality risk
& Prevents implausible absolute certainty by enforcing a documented nonzero residual when collapsing; residual size depends on correction history and corroboration strength. \\

\bottomrule
\end{tabularx}

\vspace{0.4em}

\begin{tabularx}{\textwidth}{p{0.27\textwidth} Y}
\toprule
\textbf{Reasoning Memory} & \textbf{Learned Reasoning / Calibration Patterns} \\
\midrule

Calibration experiences
& Run event-state/provenance preflight before decisive updates; use vig-removed market probabilities as the default prior; branch on high-leverage uncertainties; require provenance metadata for deterministic collapse or large moves; prohibit exact 1.0/0.0 and retain documented nonzero epsilon. \\

Overconfidence patterns
& Avoid collapsing on secondary or social reports without primary provenance; do not treat syndicated reposts as independent corroboration; do not use post-cutoff archival captures or access-time metadata as contemporaneous evidence. \\

Underconfidence patterns
& Do not maintain broad uncertainty after timestamped primary confirmations; avoid defaulting to 50/50 when a documented model or thin-market blend is available; branch when high-quality sources support a decisive scenario. \\

Probability-update lessons
& Use publisher-declared publication time or archived proof for admissibility; scale update magnitude and residual size by source tier and corroboration count; blend market-implied and model-implied priors when markets are thin or divergent; label evidence-limited forecasts and widen uncertainty when authoritative evidence is inaccessible. \\

Common reasoning failures
& Skipping event-state/provenance preflight; counting syndicated reports as independent confirmations; omitting epsilon rationale when collapsing; double-counting information already embedded in market priors. \\

\bottomrule
\end{tabularx}

\caption{Representative learned memory for the \texttt{team\_sports\_match\_winner} subcategory.}
\label{tab:memory_team_sports_match_winner}
\end{table*}
\begin{table*}[t]
\centering
\footnotesize
\setlength{\tabcolsep}{4pt}
\renewcommand{\arraystretch}{1.03}

\begin{tabularx}{\textwidth}{p{0.27\textwidth} Y}
\toprule
\multicolumn{2}{l}{\textbf{Representative Memory Example 2: \texttt{commodity\_price\_threshold\_by\_date}}} \\
\midrule
\textbf{Factor Memory} & \textbf{Typical Effect on Output} \\
\midrule

Market-implied crude-price expectation
& Sustained upward momentum shifts mass toward higher thresholds; elevated volatility increases dispersion and raises tail probabilities on both sides. \\

Refining capacity, outages, and regional throughput constraints
& Large or persistent outages increase the probability of higher thresholds, especially regionally; restored capacity lowers those probabilities. \\

Demand seasonality and short-term mobility patterns
& Rising seasonal demand increases probability of exceeding higher thresholds; falling demand or abrupt mobility drops reduce those probabilities. \\

Inventory levels and short-term stock draws/builds
& Consecutive draws increase probability of higher thresholds, while builds reduce it; inventories moderate upstream shock pass-through. \\

Institutional reporting alignment and sampling sensitivity
& Same-day authoritative postings tightly anchor the distribution; absent or differently averaged series require modeling systematic offsets and larger uncertainty. \\

Inter-series sampling and averaging differences
& Known consistent offsets shift the central estimate slightly and indicate that small gaps may be measurement differences rather than substantive moves. \\

Cross-source consensus and dispersion
& Tight cross-source clustering compresses the distribution and lowers outside-tail probabilities; broad dispersion widens uncertainty and raises tail mass. \\

Short-window tail/event penetration probability
& Credible events increase tail probabilities only if they can affect the target sampling window; otherwise tail mass remains near baseline volatility. \\

\bottomrule
\end{tabularx}

\vspace{0.4em}

\begin{tabularx}{\textwidth}{p{0.27\textwidth} Y}
\toprule
\textbf{Reasoning Memory} & \textbf{Learned Reasoning / Calibration Patterns} \\
\midrule

Calibration experiences
& Anchor tightly on authoritative same-day postings; build one continuous short-term distribution and derive threshold probabilities from it; model cent-scale inter-series offsets as measurement bias; use source clustering or dispersion to tighten or widen uncertainty. \\

Overconfidence patterns
& Avoid near-certainty from a single source without accounting for sampling or rounding; do not stack correlated corroborative signals as independent evidence; avoid large tail probability when clustered authoritative observations contradict it. \\

Underconfidence patterns
& Do not maintain broad uncertainty when same-day authoritative observations exist; avoid diluting strong late-window signals across many thresholds; downweight distal signals that cannot materially affect near-term outcomes. \\

Probability-update lessons
& Large shifts require high-reliability event-specific evidence; explicitly handle strict threshold wording and rounding; cap probabilities for outcomes contradicted by contemporaneous authoritative observations; collapse repeated reports of the same data stream into one effective update. \\

Common reasoning failures
& Confusing reporting artifacts with true price movement near cent-scale thresholds; double-counting outlets that repeat the same source; averaging target and proxy series without correcting timing or sampling differences. \\

\bottomrule
\end{tabularx}

\caption{Representative learned memory for the \texttt{commodity\_price\_threshold\_by\_date} subcategory.}
\label{tab:memory_commodity_price_threshold}
\end{table*}

\subsection{Forecasting Question Taxonomy}
Table~\ref{tab:forecast_taxonomy} presents partial forecasting question taxonomy used in our analysis on Prophet Arena dataset.
\begin{table*}[t]
\centering
\small
\setlength{\tabcolsep}{5pt}
\renewcommand{\arraystretch}{1.12}
\begin{tabular}{p{0.30\linewidth} p{0.64\linewidth}}
\toprule
\textbf{Category} & \textbf{Subcategories} \\
\midrule
Ambiguous or underspecified & Missing resolution criteria \\
\midrule
Commodity and financial prices & Commodity price threshold by date \\
\midrule
Competitive events & Individual competition winner; Playoff series winner; Team sports match winner \\
\midrule
Diplomatic meetings and encounters & Which named leader meets subject \\
\midrule
Environmental or meteorological outcomes & Global temperature index thresholds; Weather precipitation thresholds \\
\midrule
Geopolitical or foreign policy actions & Sanctions or coercive measures \\
\midrule
Government policy or executive actions & Presidential executive orders by deadline; Presidential trade policy action \\
\midrule
Legal or judicial outcomes & Arrest or charges by deadline; Lawsuit liability or verdict \\
\midrule
Legislative rules or parliamentary procedures & Filibuster threshold or rule change \\
\midrule
Legislative votes & Individual legislator vote on nomination \\
\midrule
Macroeconomic indicators & Monthly inflation index thresholds \\
\midrule
Market chart or popularity outcomes & Chart rank winner; Content engagement threshold; Content release by named artist; Critic aggregate score thresholds; Sales or streams numeric \\
\midrule
Monetary policy or central bank actions & Central bank policy rate move by date \\
\midrule
Multi-option winner markets & Candidate withdrawal or dropout by date; Corporate leadership appointment; Election margin or vote share thresholds; Political nominee or election winner; Tournament or championship winner \\
\midrule
Personal physical challenge & Individual challenge completion by date \\
\midrule
Public statements or behavior & Named individual attendance; Phrase mention; Physical action during live stream; Public appearance or sighting; Specific quote or topic \\
\midrule
Reality TV elimination & Elimination outcome \\
\midrule
Team season outcomes & Regular season win totals; Season win totals banded; Team regular season wins thresholds; Season wins thresholds; Season wins total thresholds; Season total wins prediction; Season wins threshold; Team win total thresholds \\
\bottomrule
\end{tabular}
\caption{Partial taxonomy of forecasting question categories and subcategories used in our analysis.}
\label{tab:forecast_taxonomy}
\end{table*}

\section{Detailed prompts}
\label{sec:prompts}
\subsection{Agent Prompts}
\label{sec:backbone_prompt}
\begin{table*}[h]
\centering
\small
\begin{tabular}{p{\dimexpr\linewidth-2\tabcolsep}}
\toprule
\begin{minipage}[t]{\linewidth}
\ttfamily

\textbf{Session 1: Factor Decomposition}\\

This is session 1.\\
Decompose the forecasting task into the key factors you would investigate. The max number of factors is 5, but use fewer if that seems sufficient. Focus on the most important factors that will drive the forecast.\\
If factor memory is provided, treat it as weak prior structure rather than as an authoritative template.\\
Use retrieved factor memory only to suggest candidate factors that may be relevant.\\
You may discard retrieved factors if they do not fit the current case.\\
Do not simply mirror the retrieved factor list; adapt it to the current question.\\
For each decomposed factor, also generate a short statement of its typical effect on the forecast output or probabilities.\\
For each decomposed factor, also generate a short possible error or reasoning trap that should be avoided when using that factor.\\
If factor memory is provided, use it as background context, but do not copy or manually select an error pattern from it. Infer the most relevant error-to-avoid yourself.\\
Factor memory may help rank or surface plausible drivers, but it must not by itself determine the final decomposition or imply strong confidence.\\
Do not use any tools.\\
Do not give a final answer or probabilities yet.\\
Return only valid JSON with a top-level key `factors'.\\
`factors' must be a list of objects with keys `factor\_name', `rationale', `typical\_effect\_on\_output', and `potential\_common\_error\_pattern'.\\
Set `typical\_effect\_on\_output' to a concise description of how the factor usually shifts probabilities or changes the forecast.\\
Set `potential\_common\_error\_pattern' to a concise possible error to avoid for that factor, or an empty string only if none is relevant.\\
Use short reusable factor names rather than full-sentence descriptions.\\

\vspace{4pt}
\textbf{Session 2: Evidence Gathering and Forecasting}\\
This is session 2.\\
Use the prior factor decomposition from the conversation as your search plan.\\
If reasoning memory is provided, use it as a reusable prior rather than as ground truth.\\
Treat the retrieved reasoning patterns as explicit guardrails for how to reason about the task.\\
You should actively use them to avoid the listed common errors, distribution mistakes, calibration mistakes, and other reasoning traps.\\
You CAN ONLY use web search at most \{max\_search\_calls\} times total.\\
You MUST finish the task within \{max\_turns\} turns total for this session.\\
Budget your turns carefully and keep enough remaining turns to produce the final answer.\\
If the evidence is already sufficient, stop searching early rather than risking a max-turns failure.\\
Do not spend your last available turn on search or extra reasoning; reserve it for the final JSON answer.\\
Do not use all searches unless necessary.\\
Once you have enough evidence, stop searching and return the final JSON immediately.\\
If reasoning memory is provided in the user input, use it as a checklist for reasoning and evidence gathering, and explicitly check your forecast against it before answering.\\

\end{minipage} \\
\bottomrule
\end{tabular}
\caption{Prompt templates for the two-session forecasting agent. Session 1 decomposes the forecasting task into predictive factors, while Session 2 uses the factor decomposition and reasoning memory for evidence gathering and probabilistic forecasting.}
\label{prompt:two_session_forecasting_agent}
\end{table*}

\subsection{Taxonomy-relevant prompt.}
\paragraph{Taxonomy classification prompt}
We use the Prompt in Table~\ref{prompt:taxonomy_match} to determine whether each forecasting question can be assigned to an existing taxonomy entry. Given the current taxonomy and a new question, the model either matches the question to an existing category--subcategory pair or proposes a new category and subcategory when no suitable match exists. The output is constrained to a fixed JSON schema to ensure consistent downstream taxonomy updates.

\paragraph{Taxonomy revision prompt.}
We use the prompt in Table~\ref{prompt:taxonomy_promotion} to decide whether grouped unmatched questions should be incorporated into the forecasting taxonomy. Given the existing taxonomy and a grouped proposal, the model determines whether the proposal is already covered, should be added as a new subcategory under an existing category, or requires a new top-level category. The output follows a fixed JSON schema to support controlled and consistent taxonomy expansion.
\begin{table*}[h]
\centering
\small
\begin{tabular}{p{\dimexpr\linewidth-2\tabcolsep}}
\toprule
\begin{minipage}[t]{\linewidth}
\ttfamily

Taxonomy:\\
\{taxonomy\_text\}\\

Question:\\
\{question\}\\

Return a JSON object with these keys:\\
- matched\_existing\_taxonomy: boolean\\
- category\_name: string or null\\
- subcategory\_name: string or null\\
- confidence: number between 0 and 1\\
- rationale: short string\\
- proposed\_category\_name: string or null\\
- proposed\_subcategory\_name: string or null\\
- proposed\_category\_rationale: string or null\\

\textbf{Rules:}\\
- If the question fits an existing category and subcategory, set matched\_existing\_taxonomy=true and fill category\_name/subcategory\_name.\\
- If it does not fit, set matched\_existing\_taxonomy=false and propose a new category and subcategory.\\
- Do not invent multiple options.\\
- Keep rationale concise.\\

\end{minipage} \\
\bottomrule
\end{tabular}
\caption{Prompt template for matching a forecasting question to an existing taxonomy or proposing a new taxonomy entry.}
\label{prompt:taxonomy_match}
\end{table*}
\begin{table*}[h]
\centering
\small
\begin{tabular}{p{\dimexpr\linewidth-2\tabcolsep}}
\toprule
\begin{minipage}[t]{\linewidth}
\ttfamily

Existing taxonomy:\\
\{taxonomy\_text\}\\

Grouped proposed new category candidate:\\
\{proposal\_json\}\\

Decide whether this grouped proposal should be promoted into the taxonomy.\\

Return a JSON object with these keys:\\
- promote\_to\_taxonomy: boolean\\
- target\_category\_name: string or null\\
- target\_subcategory\_name: string or null\\
- create\_new\_category: boolean\\
- create\_new\_subcategory: boolean\\
- rationale: string\\
- category\_summarized\_patterns: array of strings\\
- subcategory\_summarized\_patterns: array of strings\\
- examples: array of strings\\

\textbf{Rules:}\\
- If the proposal is actually covered by an existing category/subcategory, do not promote it as new.\\
- If it should map to an existing category but needs a new subcategory, set create\_new\_category=false and create\_new\_subcategory=true.\\
- If it needs a genuinely new top-level category, set create\_new\_category=true and create\_new\_subcategory=true.\\
- Use concise snake\_case names for category and subcategory.\\
- Keep at most 3 summarized patterns and at most 3 examples.\\
- Examples should be selected from the proposal examples when possible.\\
- Always return valid JSON only.\\

\end{minipage} \\
\bottomrule
\end{tabular}
\caption{Prompt template for deciding whether grouped proposed taxonomy candidates should be promoted into the forecasting taxonomy.}
\label{prompt:taxonomy_promotion}
\end{table*}

\subsection{Memory-relevant prompts}
\paragraph{Subcategory memory revision suggestion prompt.}
We use the prompt in Table~\ref{prompt:subcategory_memory_revision} to propose the  revision for the subcategory-level forecasting memory from paired event trajectories. For each event, the model compares the filtered prediction under realistic forecasting conditions with a no-filter reference and the ground truth, then identifies which parts of the existing memory were useful, missing, overly specific, or misleading. The prompt explicitly separates factor-memory revisions from reasoning and calibration revisions, ensuring that predictive dimensions are updated independently from probability-update rules or confidence-control heuristics.
\begin{table*}[t]
\centering
\small
\begin{tabular}{p{\dimexpr\linewidth-2\tabcolsep}}
\toprule
\begin{minipage}[t]{\linewidth}
\ttfamily

You are reviewing one forecasting event to suggest how a subcategory memory should be revised.\\

Category: \{category\_name\}\\
Subcategory: \{subcategory\_name\}\\
Existing memory: \{existing\_memory\_json\}\\

You are given two bundles for the same event:\\
1. FILTERED bundle: inference under a date-based search cutoff.\\
2. NO-FILTER bundle: inference without date restriction, used as an oracle reference.\\

Filtered event bundle: \{filtered\_event\_bundle\}\\
No-filter reference bundle: \{nofilter\_event\_bundle\}\\
Ground truth: \{ground\_truth\}\\

\textbf{Task:}\\
Compare the filtered inference with the no-filter reference and ground truth. Identify which parts of the existing memory were useful, missing, too concrete, or misleading. Suggest reusable subcategory-level revisions, not event-specific facts. Separate factor-memory issues from reasoning and calibration issues.\\
\medskip
\textbf{Return valid JSON only with:}\\
- category\_name: string\\
- subcategory\_name: string\\
- memory\_strengths: array of strings, at most 5\\
- memory\_gaps: array of strings, at most 5\\
- suggested\_factor\_revisions: array of objects, at most \{max\_factors\}, each with:\\
\hspace*{1.5em}- factor\_name: string\\
\hspace*{1.5em}- action: one of [``keep'', ``revise'', ``add'', ``deprioritize'']\\
\hspace*{1.5em}- description: string\\
\hspace*{1.5em}- typical\_effect\_on\_output: string\\
\hspace*{1.5em}- factor\_specific\_checks: array of 2--4 short strings\\
\hspace*{1.5em}- common\_failures: array of 2--4 short strings\\
- suggested\_calibration\_revisions: array of strings, at most 6\\
- suggested\_reasoning\_failure\_revisions: array of strings, at most 6\\
- event\_specific\_notes: string\\
\medskip
\textbf{Rules:}\\
Factor memory describes broad predictive dimensions that matter in this subcategory. Reasoning memory describes how to update probabilities, calibrate confidence, interpret evidence, and handle uncertainty. Each factor revision must be a broad latent predictive dimension, not a named entity, search query, raw metric, threshold, confidence cap, provenance rule, or procedural update rule. Put calibration rules, confidence-control policies, late-window verification rules, and evidence-strength thresholds into calibration or reasoning revisions instead.\\

Suggested factors should remain meaningful even if numeric update rules are removed. Good factor names resemble market-implied expectation, structural opportunity and role, institutional signaling, macro trend alignment, timing feasibility, or electorate alignment. Avoid factor names such as evidence credibility, provenance, update caps, thresholds, historical priors, regularization, or late-window verification cadence.\\

Focus especially on calibration: distinguish evidence that only helps rank outcomes from evidence strong enough to justify extreme confidence. If the no-filter reference also fails to approach ground truth, note that the missing signal may be genuinely hard to capture rather than a clear memory gap.\\

Before finalizing each suggested factor revision, check whether it is truly a predictive dimension. If it is mainly a reasoning or calibration rule, move it to suggested\_calibration\_revisions or suggested\_reasoning\_failure\_revisions.\\
\end{minipage} \\
\bottomrule
\end{tabular}
\caption{Prompt template for revising subcategory-level forecasting memory from filtered and no-filter event bundles.}
\label{prompt:subcategory_memory_revision}
\end{table*}

\paragraph{Batch memory-suggestion summarization prompt.}
We use the prompt in Table~\ref{prompt:batch_memory_summary} to aggregate per-event memory revision suggestions within the same forecasting subcategory. The model identifies revision signals that recur across multiple events and summarizes them into reusable factor, calibration, and reasoning-failure lessons. One-off event-specific observations are discarded so that the resulting summary can be reliably merged with other batch summaries before the final memory update.
\begin{table*}[t]
\centering
\small
\begin{tabular}{p{\dimexpr\linewidth-2\tabcolsep}}
\toprule
\begin{minipage}[t]{\linewidth}
\ttfamily

You are summarizing a batch of per-event memory revision suggestions for one forecasting subcategory.\\

Category: \{category\_name\}\\
Subcategory: \{subcategory\_name\}\\
Existing memory, built from \{n\_existing\_questions\} questions: \{existing\_memory\_json\}\\
Per-event revision suggestions for this batch, \{batch\_size\} events: \{event\_suggestions\_json\}\\

\textbf{Task:}\\
Identify signals that recur across multiple events in this batch. Summarize recurring factor revision signals, calibration lessons, and reasoning failures. Discard one-off, event-specific observations that appear in only one event. Be concise, since this summary will be combined with summaries from other batches before final memory revision.\\

\textbf{Return valid JSON only with:}\\
- batch\_size: integer\\
- recurring\_factor\_signals: array of objects, at most \{max\_factors\}, each with:\\
\hspace*{1.5em}- factor\_name: string\\
\hspace*{1.5em}- action: one of [``keep'', ``revise'', ``add'', ``deprioritize'']\\
\hspace*{1.5em}- frequency: integer\\
\hspace*{1.5em}- consensus\_description: string\\
\hspace*{1.5em}- key\_checks: array of strings, at most 3\\
\hspace*{1.5em}- key\_failures: array of strings, at most 3\\
- recurring\_calibration\_signals: array of strings, at most 6\\
- recurring\_reasoning\_failure\_signals: array of strings, at most 6\\
- common\_memory\_gaps: array of strings, at most 4\\
- common\_memory\_strengths: array of strings, at most 4\\

\textbf{Rules:}\\
Only include factor, calibration, and reasoning-failure signals supported by at least two events in the batch. Omit one-off event-specific observations. Keep all descriptions short, reusable, and independent of individual event details.\\

\end{minipage} \\
\bottomrule
\end{tabular}
\caption{Prompt template for summarizing recurring memory-revision signals across a batch of forecasting events.}
\label{prompt:batch_memory_summary}
\end{table*}

\paragraph{Final subcategory memory revision prompt.}
We use the prompt in table~\ref{prompt:final_subcategory_memory_revision} to produce the final memory update for each forecasting subcategory. The model synthesizes batch-level revision summaries from the current epoch and updates the existing memory only when changes are supported by repeated evidence across events. The prompt enforces conservative factor preservation, separates predictive factors from reasoning and calibration rules, and outputs a structured JSON memory used in later forecasting episodes.
\begin{table*}[t]
\centering
\resizebox{\textwidth}{!}{
\begin{tabular}{p{0.98\textwidth}}
\toprule
\begin{minipage}[t]{0.98\textwidth}
\scriptsize
\ttfamily

You are revising reusable forecasting memory for one subcategory.\\

Category: \{category\_name\}\\
Subcategory: \{subcategory\_name\}\\
Existing memory, built from \{n\_existing\_questions\} questions: \{existing\_memory\_json\}\\
Batch-merged revision summaries from the current epoch, \{n\_new\_questions\} new questions: \{event\_revision\_suggestions\_json\}\\

Each suggestion was produced by comparing filtered inference with a no-filter oracle reference and ground truth.\\

\textbf{Task:}\\
Synthesize all revision summaries. Revise memory only when supported by multiple events. Keep factors reusable, high-level, and stable. Preserve useful existing content unless repeated evidence supports revision.\\

\textbf{Factor preservation rules:}\\
Every existing factor must be explicitly addressed; KEEP is the default. Modify a factor only if at least one suggestion explicitly targets it with action=``revise''. Drop a factor only if at least two suggestions deprioritize or contradict it and it is not independently useful. If \{n\_new\_questions\} $<$ \{n\_existing\_questions\}, prefer extending or merging over replacing. Add a factor only if it captures a genuinely absent predictive dimension. Silent replacement by a renamed equivalent factor is not allowed.\\

\textbf{Return valid JSON only with:}\\
- category\_name: string\\
- subcategory\_name: string\\
- revised\_common\_factors: array of objects, at most \{max\_factors\}, each with factor\_name, description, typical\_effect\_on\_output, factor\_specific\_checks, and common\_failures\\
- revised\_common\_reasoning\_patterns: object with calibration\_experiences, overconfidence\_patterns, underconfidence\_patterns, probability\_update\_lessons, and common\_reasoning\_failures\\
- revised\_representative\_examples: array of strings, at most 5\\
- revised\_notes: string, 1--3 sentences\\
- update\_rationale: string, 1--3 sentences\\

\textbf{Core separation:}\\
Factor memory describes broad predictive dimensions. Reasoning memory describes probability updating, calibration, confidence control, evidence interpretation, and uncertainty handling. Calibration rules, confidence caps, provenance rules, thresholds, regularization policies, and late-window verification rules must be placed in revised\_common\_reasoning\_patterns, not revised\_common\_factors.\\

\textbf{Revision rules:}\\
Preserve useful content by default; merge, refine, or extend rather than drop unless repeated evidence supports removal. Prioritize stable cross-event patterns over one-off anomalies. Each revised factor must be a broad latent predictive dimension, not a named entity, search query, raw metric, threshold, confidence cap, or procedural update rule. Good factors resemble market-implied expectation, structural opportunity and role, institutional signaling, macro trend alignment, timing feasibility, or electorate alignment.\\

\textbf{Reasoning rules:}\\
Store cross-factor lessons in revised\_common\_reasoning\_patterns. Focus on calibration: when to stay near base rates, when weak signals should not be stacked, when large probability shifts require direct high-reliability evidence, and how to handle sparse, conflicting, or uncertain evidence. Distinguish ranking evidence from evidence strong enough for extreme confidence.\\

\textbf{Final check:}\\
Before finalizing each revised factor, check whether it is truly a predictive dimension. If it is mainly a reasoning or calibration rule, move it to revised\_common\_reasoning\_patterns.\\

\end{minipage} \\
\bottomrule
\end{tabular}
}
\caption{Prompt template for final subcategory-level forecasting memory revision.}
\label{prompt:final_subcategory_memory_revision}
\end{table*}

\paragraph{Memory artifact evaluation prompt.}
\label{app:prompt_mem_eval}
We use the following prompt to evaluate the quality of generated memory artifacts. Each artifact is scored along five dimensions: generalizability, novelty, completeness, accuracy, and actionability. The evaluator is instructed to return a fixed JSON object with integer scores and a one-sentence rationale.
\begin{table*}[h]
\centering
\small
\begin{tabular}{p{\dimexpr\linewidth-2\tabcolsep}}
\toprule
\begin{minipage}[t]{\linewidth}
\ttfamily

You are an expert evaluator of AI-generated memory artifacts for a probabilistic forecasting system. Be concise, critical, and accurate.\\

A memory artifact is a stored knowledge unit that an AI forecasting agent retrieves to improve future probability estimates.\\

Rate the following artifact on 5 dimensions, each scored 1--10:\\

1. generalizability -- applies broadly across question types, not overfit to specific examples\\
2. novelty -- contains non-obvious insights beyond common knowledge\\
3. completeness -- covers the key factors needed for forecasting this question type\\
4. accuracy -- factually correct and based on sound reasoning\\
5. actionability -- directly guides forecasting decisions, e.g., ``look for X'' or ``weight Y''\\

Memory artifact:\\
---\\
\{memory\_text\}\\
---\\

Reply ONLY with valid JSON, with no markdown fences:\\
\{\{``generalizability'': $<$int$>$, ``novelty'': $<$int$>$, ``completeness'': $<$int$>$, ``accuracy'': $<$int$>$, ``actionability'': $<$int$>$, ``reasoning'': ``$<$one sentence$>$''\}\}\\

\end{minipage} \\
\bottomrule
\end{tabular}
\caption{Prompt template for evaluating generated memory artifacts across five quality dimensions.}
\label{prompt:memory_artifact_eval}
\end{table*}

\section{Limitations and Social Impact}
\paragraph{Limitations}
\label{limitation}
This work studies memory-augmented agentic forecasting under a chronological evaluation protocol. There remain challenges in interpreting and verifying stored memories, especially when memories are abstracted from many prior experiences. Future work could develop more transparent memory inspection and validation mechanisms for forecasting agents.

\paragraph{Societal Impact.}
\label{impact}
Automated forecasting systems may create risks if users over-rely on their predictions, or if forecasts are used in high-stakes settings without appropriate human oversight. Forecasting agents should therefore be deployed with transparency about uncertainty, careful evaluation in the target domain, and safeguards against using generated probabilities as sole decision criteria.



\clearpage
\section*{NeurIPS Paper Checklist}

\begin{enumerate}

\item {\bf Claims}
    \item[] Question: Do the main claims made in the abstract and introduction accurately reflect the paper's contributions and scope?
    \item[] Answer: \answerYes{} 
    \item[] Justification: The abstract and introduction accurately summarize the paper's scope and contributions

\item {\bf Limitations}
    \item[] Question: Does the paper discuss the limitations of the work performed by the authors?
    \item[] Answer: \answerYes{} 
    \item[] Justification: The limitation section is in Appendix~\ref{limitation}.

\item {\bf Theory assumptions and proofs}
    \item[] Question: For each theoretical result, does the paper provide the full set of assumptions and a complete (and correct) proof?
    \item[] Answer: \answerNA{} 
    \item[] Justification: The paper does not include theoretical results. 

    \item {\bf Experimental result reproducibility}
    \item[] Question: Does the paper fully disclose all the information needed to reproduce the main experimental results of the paper to the extent that it affects the main claims and/or conclusions of the paper (regardless of whether the code and data are provided or not)?
    \item[] Answer: \answerYes{} 
    \item[] Justification: The detailed experimental settings are provided in Appendix~\ref{app:setup}, and prompts are shown in Appendix~\ref{sec:prompts}.

\item {\bf Open access to data and code}
    \item[] Question: Does the paper provide open access to the data and code, with sufficient instructions to faithfully reproduce the main experimental results, as described in supplemental material?
    \item[] Answer: \answerNo{} 
    \item[] Justification: We will release full code when the paper is accepted.

\item {\bf Experimental setting/details}
    \item[] Question: Does the paper specify all the training and test details (e.g., data splits, hyperparameters, how they were chosen, type of optimizer) necessary to understand the results?
    \item[] Answer: \answerYes{}{} 
    \item[] Justification: The detailed experimental settings are provided in Appendix~\ref{app:setup}, and prompts are shown in Appendix~\ref{sec:prompts}.

\item {\bf Experiment statistical significance}
    \item[] Question: Does the paper report error bars suitably and correctly defined or other appropriate information about the statistical significance of the experiments?
    \item[] Answer: \answerYes{} 
    \item[] Justification: The robustness of our method is shown in Appendix~\ref{app:bootstrap_ci}.

\item {\bf Experiments compute resources}
    \item[] Question: For each experiment, does the paper provide sufficient information on the computer resources (type of compute workers, memory, time of execution) needed to reproduce the experiments?
    \item[] Answer: \answerYes{} 
    \item[] Justification: The paper reports the model backbones, inference settings, memory-revision epochs, and token usage. Since the experiments use API-based LLM inference rather than local training, token counts and LLM call counts serve as the primary measure of computational cost.
    
\item {\bf Code of ethics}
    \item[] Question: Does the research conducted in the paper conform, in every respect, with the NeurIPS Code of Ethics \url{https://neurips.cc/public/EthicsGuidelines}?
    \item[] Answer: \answerYes{} 
    \item[] Justification: We follow the NeurIPS Code of821
Ethics.

\item {\bf Broader impacts}
    \item[] Question: Does the paper discuss both potential positive societal impacts and negative societal impacts of the work performed?
    \item[] Answer: \answerYes{} 
    \item[] Justification: The social impact is discussed in Appendix~\ref{impact}
    
\item {\bf Safeguards}
    \item[] Question: Does the paper describe safeguards that have been put in place for responsible release of data or models that have a high risk for misuse (e.g., pre-trained language models, image generators, or scraped datasets)?
    \item[] Answer: \answerNA{} 
    \item[] Justification: The paper does not release high-risk data or models, such as pretrained language models, image generators, or scraped datasets. The experiments use existing open-source forecasting datasets and model APIs, and we do not redistribute proprietary model weights or newly collected high-risk data.

\item {\bf Licenses for existing assets}
    \item[] Question: Are the creators or original owners of assets (e.g., code, data, models), used in the paper, properly credited and are the license and terms of use explicitly mentioned and properly respected?
    \item[] Answer: \answerYes{} 
    \item[] Justification: The forecasting datasets used in this paper are open-source and are cited in the experimental setup.

\item {\bf New assets}
    \item[] Question: Are new assets introduced in the paper well documented and is the documentation provided alongside the assets?
    \item[] Answer: \answerNA{} 
    \item[] Justification: The paper does not release new assets.

\item {\bf Crowdsourcing and research with human subjects}
    \item[] Question: For crowdsourcing experiments and research with human subjects, does the paper include the full text of instructions given to participants and screenshots, if applicable, as well as details about compensation (if any)? 
    \item[] Answer: \answerNA{} 
    \item[] Justification: The paper does not involve crowdsourcing nor research with human subjects.

\item {\bf Institutional review board (IRB) approvals or equivalent for research with human subjects}
    \item[] Question: Does the paper describe potential risks incurred by study participants, whether such risks were disclosed to the subjects, and whether Institutional Review Board (IRB) approvals (or an equivalent approval/review based on the requirements of your country or institution) were obtained?
    \item[] Answer: \answerNA{} 
    \item[] Justification: The paper does not involve crowdsourcing nor research with human subjects.

\item {\bf Declaration of LLM usage}
    \item[] Question: Does the paper describe the usage of LLMs if it is an important, original, or non-standard component of the core methods in this research? Note that if the LLM is used only for writing, editing, or formatting purposes and does \emph{not} impact the core methodology, scientific rigor, or originality of the research, declaration is not required.
    \item[] Answer: \answerYes{} 
    \item[] Justification: LLMs are used as core components of the method. The paper reports the backbone models and experimental setup, and provides the prompts for the LLM-based taxonomy and memory-revision modules in the appendix.

\end{enumerate}

\end{document}